\pdfoutput=1

\documentclass[11pt]{article}
\usepackage{tabularx}
\usepackage{arydshln} 
\usepackage[dvipsnames]{xcolor}

\usepackage[preprint]{acl}
\usepackage{xltabular}

\usepackage{times}
\usepackage{latexsym}
\usepackage{amsfonts}

\usepackage[T1]{fontenc}

\usepackage[utf8]{inputenc}

\usepackage{microtype}

\usepackage{inconsolata}

\usepackage{graphicx}

\usepackage{lipsum} 
\usepackage{comment}

\usepackage{longtable}
\usepackage{makecell}
\usepackage{array}
\usepackage{booktabs, colortbl}
\usepackage{marginnote} 

\usepackage[normalem]{ulem} 
\title{From Tools to Teammates: Evaluating LLMs\\in Multi-Session Coding Interactions}

\author{
 \textbf{Nathanaël Carraz Rakotonirina\textsuperscript{1}},
 \textbf{Mohammed Hamdy\textsuperscript{5}},
 \textbf{Jon Ander Campos\textsuperscript{3}},
 \textbf{Lucas Weber\textsuperscript{1}},
\\
 \textbf{Alberto Testoni\textsuperscript{6}},
 \textbf{Marzieh Fadaee\textsuperscript{4}},
 \textbf{Sandro Pezzelle\textsuperscript{2}},
 \textbf{Marco Del Tredici}
\\
\\
 \textsuperscript{1}Universitat Pompeu Fabra,
  \textsuperscript{2}University of Amsterdam,
 \textsuperscript{3}Cohere, \\
 \textsuperscript{4}Cohere Labs,
 \textsuperscript{5}Cohere Labs Community,
 \textsuperscript{6}Amsterdam UMC
}

\begin{document}
\maketitle
\begin{abstract}

\let\thefootnote\relax\footnotetext{Correspondence:  \href{mailto:nathanael.rakotonirina@upf.edu}
{nathanael.rakotonirina@upf.edu} \\
Marco Del Tredici started this project while working at Cohere and Alberto Testoni while working at the University of Amsterdam. Lucas Weber is now at Fraunhofer IIS.
}

Large Language Models (LLMs) are increasingly used in  working environments for a wide range of tasks, excelling at solving individual problems in isolation. However, are they also able to effectively collaborate over long-term interactions? To investigate this, we introduce \textbf{\textsc{MemoryCode}}, a synthetic multi-session dataset designed to test LLMs' ability to track and execute simple coding instructions amid irrelevant information, simulating a realistic setting. 
While all the models we tested handle isolated instructions well, even state-of-the-art models like GPT-4o and the reasoning model DeepSeek-R1 show degraded performance when instructions are spread across sessions. Our analysis suggests this is due to their failure to retrieve and integrate information over long instruction chains. Our results highlight a fundamental limitation of current LLMs, restricting their ability to collaborate effectively in long interactions\footnote{The code and data are available at \url{https://github.com/Cohere-Labs-Community/MemoryCode}.}.

\end{abstract}

\section{Introduction}
\label{sec:intro}

Current efforts to improve the performance of large language models (LLMs) mostly focus on their ability to solve increasingly harder tasks autonomously. Examples of this research include solving complex math~\cite{wangmathcoder,gao2024omni,alphageo2021}, coding~\cite{chen2021evaluating,austin2021program,tao2024crystal,puerto-etal-2024-code}, or reasoning problems~\cite{hao2023reasoning,wang2024chain,renze2024self}.
Since many of these tasks are relevant to real-world applications, LLMs are widely adopted in industry, where they have been reported to significantly enhance productivity~\cite{weber2024significant,cambon2023early}.
This extensive adoption of LLM assistants into the daily working routine is effectively converting them from mere tools to fully-fledged \textit{teammates}. 
For LLMs to behave as such, though, complementary skills related to collaboration and interaction are needed. 
One such ability is retaining relevant information from multiple interactions with human users
and leveraging it for future tasks.

\begin{figure}[t]
    \centering
    
    \includegraphics[width=1\columnwidth]{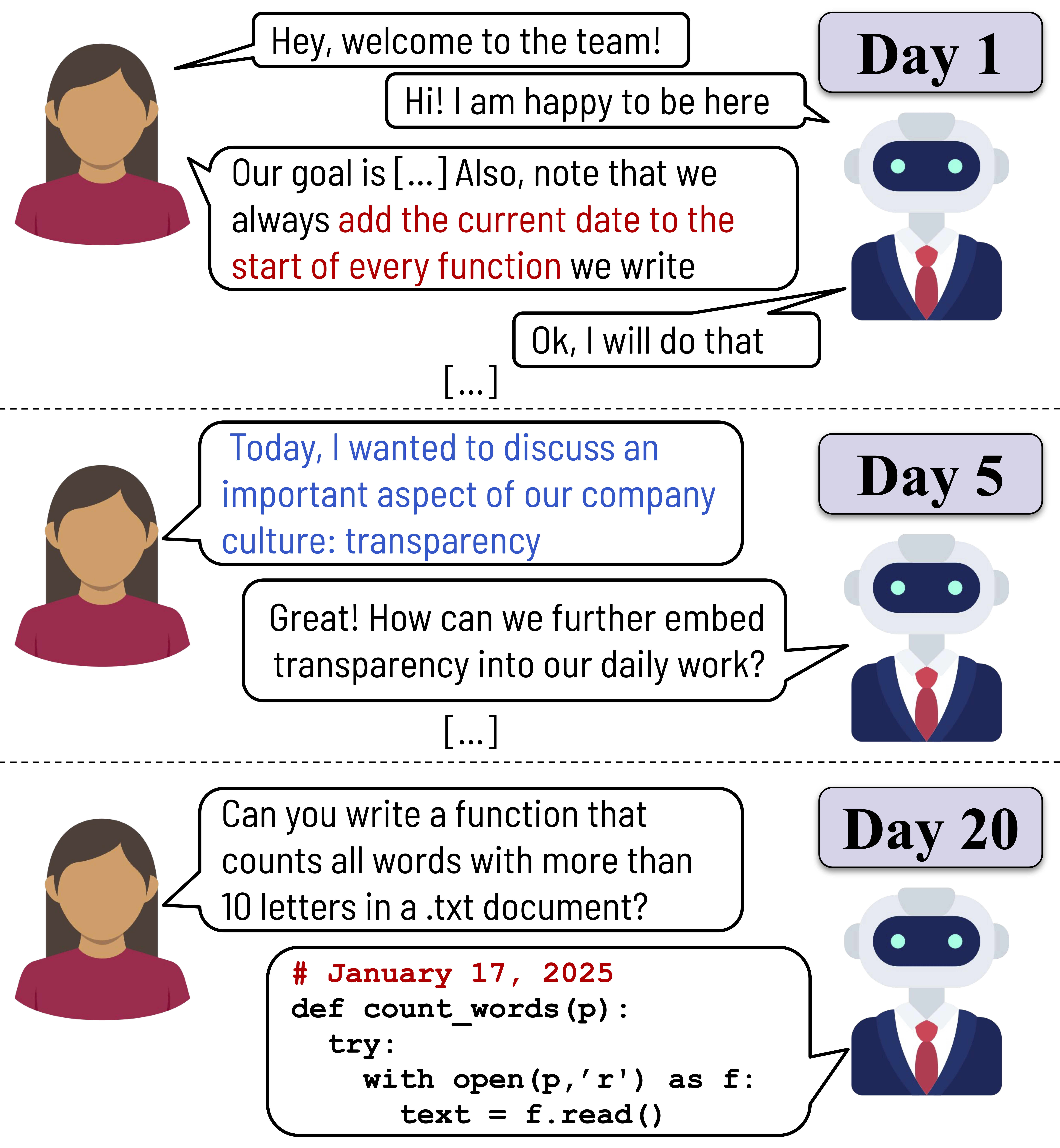}
    \caption{A simplified but realistic example of a long-term interaction between a human and an LLM-based `teammate'. 
    In this example, each day represents a single session. The LLM teammate must remember a piece of information---in red---learned during the session on Day 1 to correctly perform a task on Day 20, while also receiving irrelevant information---in blue---on Day 5.
    }
    \label{fig:memoryCode_fig1_sketch}
\end{figure}

In this paper, we investigate this challenge by introducing \textbf{\textsc{MemoryCode}}, 
a synthetic dataset 
of multi-session dialogue histories 
designed to evaluate models' ability to 
track simple coding instructions provided amid irrelevant information, and execute them in future coding tasks.
Each dialogue history is a chronological sequence of dialogues, or \textit{sessions}, between a \textit{mentor} and a \textit{mentee}.
Throughout the sessions, the mentor passes critical information for solving a task to a mentee. Crucially, this information is interspersed with a substantial amount of unrelated content, thus reflecting the real-life scenario of working in an office. 
Furthermore, the information needed to perform a task can be updated multiple times throughout the dialogue history.

\textsc{MemoryCode} mimics natural interactions between coworkers. Figure \ref{fig:memoryCode_fig1_sketch} shows an example of such interactions, where various coding conventions and rules arise~\cite{convertino2008articulating,chumg2022drives} that are passed on to new team members (Day 1), often among other pieces of information irrelevant to coding tasks (Day 5). Newcomers are expected to comply with such rules when performing future tasks (Day 20), unless rules are deprecated or changed.
\textsc{MemoryCode} tests whether current models behave like new human teammates by consistently adhering to such rules across many sessions.

Similar to previous work~\cite{nelson2024needle,epstein2024mmmt,maharana-etal-2024-evaluating}, the primary goal of our benchmark is to retrieve important information from a long conversational history.  
In contrast to previous datasets, \textsc{MemoryCode} requires to \emph{use} retrieved information in practical tasks while not being explicitly cued to do so.
This is more challenging than cued retrieval of static information, as it requires prospective 
memory and spontaneous retrieval~\cite{mcdaniel2007prospective,brandimonte2014prospective}. 
Additionally, \textsc{MemoryCode} requires an integration of information retrieved from different parts of the dialogue history, as rules can be updated, with only the last update being eventually relevant.
At the same time, the rules in it (e.g., adding a date at the start of every new code) are simple to execute, which allows for disentangling a model's retrieval capabilities from other complex skills.
To the best of our knowledge, \textsc{MemoryCode} is the first multi-session dataset that tests this practically highly relevant skill.

We test several proprietary and open-source
SotA models on \textsc{MemoryCode}, and show that:
(i) Even small models succeed in executing the single coding instructions in \textsc{MemoryCode} when prompted without additional complex context, indicating that such instructions are well within the reach of current LLMs;
(ii) As we increase the complexity and provide a full mentor-mentee session, including several instructions and irrelevant information, 
only larger models continue to perform well, while the performance of smaller models drops significantly;
(iii)When we provide the full dialogue history, even strong proprietary models struggle to follow our simple instructions, with GPT-4o showing a dramatic 67\% drop in accuracy compared to its performance with instructions alone. This reveals that \textsc{MemoryCode} is a challenging benchmark even for the best available models, that struggle to retrieve and incrementally update relevant information.

We argue that solving \textsc{MemoryCode} requires more than simply scaling models even further. 
Instead, our results indicate a pressing need to develop dedicated mechanisms to enhance LLMs' abilities, such as improved long-term memory retention strategies, prospective memory, or additional reasoning mechanisms. We release the dataset under the Apache 2.0 license.

\section{Related work}
\label{sec:related}

\subsection{Long-Context Evaluation} 
Early approaches to evaluating long-context understanding date back to the pre-LLM era.
One such example is 
LAMBADA \citep{paperno-etal-2016-lambada}, which includes high-quality human-annotated samples with an average length of 75 tokens. 
As context lengths increased, new datasets were created by repurposing or expanding existing NLP datasets \citep{an-etal-2024-l, bai-etal-2024-longbench, dong-etal-2024-bamboo}. 
More recently, controlled-length synthetic evaluation frameworks, such as Needle-in-a-Haystack \cite{kamradt2023needle} and LTM \cite{LMT}, have been widely adopted for evaluating long-context understanding \citep{anil2023gemini, anthropic2024claude3}. 
In these frameworks, the %
models are
tasked with retrieving information from long distractor texts.
RULER \citep{hsieh2024ruler} extends Needle-in-the-Haystack by varying the types and numbers of \textit{needles} and adding new tasks like variable tracking and frequent word extraction.
LOFT \citep{lee2024can} adds many real-world tasks, such as Retrieval-Augmented Generation and SQL-like tasks, that require context up to millions of tokens. 
Similar to these approaches, in this work we evaluate long-context understanding in conversational settings. Unlike
other works, though,
we do not ask the models to retrieve a piece of information, but rather challenge them to retrieve 
the most up-to-date instructions dispersed across the dialogue history to accomplish a task.\\

\subsection{Long-term Dialogue Evaluation}
Multi-turn and multi-session interactions are the \textit{de facto} standard setup in which LLMs are used. Accordingly, several datasets have been introduced to evaluate long contexts in conversations.
\citealp{zheng2024judging} introduced MT-Bench, a high-quality, multi-turn question dataset across 8 knowledge categories, but with only two turns per session. 
Many benchmarks have been proposed to expand or improve upon MT-Bench \citep{sun2024parrot, bai-etal-2024-mt, kwan2024mt}. 
For example, MT-Eval \citep{kwan2024mt} evaluates different aspects of multi-turn dialogue such as the ability to understand follow-up questions.
MINT \citep{wang2024mint} focuses on tool use and natural language feedback evaluation, while \citealp{duan-etal-2024-botchat} introduce a framework where three different evaluation strategies are proposed: evaluating each multi-turn dialogue separately, comparing the quality of two generated dialogues, and comparing two dialogues to determine which one is 
the human
conversation.

Most similar to our work, MMMT-IF \citep{epstein2024mmmt} extends multi-turn and multi-modal datasets to measure instruction-following abilities and shows that the main challenge for LLMs is not in following instructions, but rather in retrieving it. Unlike MMMT-IF, our dataset was generated entirely from scratch and is specifically designed for real-world conversational settings that feature constantly evolving instructions.
In multi-session dialogues, many datasets were also created synthetically; for example, Conversation Chronicles~\citep{jang2023conversation}, which includes 200K conversations of about 5 sessions each, or LoCoMo~\citep{maharana-etal-2024-evaluating}, a multi-modal dataset based on a framework that leverages personas and temporal event graphs.
\citet{kim2024share} and \citet{kim2024dialsim} utilized movie scripts to construct complex multi-session dialogue datasets having, for example, multi-party conversations and shared memories between speakers. 
Most of the works mentioned above focus on expanding the number of turns and sessions or on introducing more complex tasks that are challenging for LLMs. Similarly, our work evaluates the performance of LLMs in multi-turn/session dialogues, but with very simple tasks and more focus on practical, real-world settings in which information is constantly changing.

\subsection{Synthetic Dialogue Generation}
Synthetic data generation via LLMs addresses limitations of human-based dataset construction such as high costs~\citep{doi:10.1073/pnas.2305016120} and privacy concerns~\citep{kurakin2023harnessing}. Precisely because of these advantages, we decided to adopt synthetic generation for the creation of \textsc{MemoryCode}. Examples of widely adopted synthetic datasets include SODA \cite{kim-etal-2023-soda}, an open-domain dialogue dataset grounded on commonsense knowledge, containing millions of utterances generated by GPT3.5;
DialHalu~\citep{chen2024diahalu}, a dataset to evaluate different subtypes of hallucination in language models; and 
MoralDial \citep{sun-etal-2023-moraldial}, which evaluates moral values in language models. 
\citet{wu2024hiding} proposed a dialogue generation framework that provides control over many attributes of the speakers, such as personality, age group, and profession. Finally, \citet{rakotonirina-baroni-2024-memoryprompt} introduced a synthetic dataset consisting of sequences of realistic facts that may be updated over time. Their dataset is designed to evaluate LLMs' ability to track specific pieces of information amid distractors. 
While similar in spirit to these approaches, our dataset is novel as it is composed of interactions set in practical business contexts and with a focus on coding.
Additionally, our evaluation emphasizes the model's ability to follow well-defined instructions rather than simply retrieving facts.

\section{Dataset}
\label{sec:data}
We simulate a scenario in which the model assumes the role of a new hire (henceforth, the \textit{mentee}) who undergoes an onboarding process in a given company. 
The mentee interacts with a \textit{mentor} in chronologically ordered \textit{sessions}. A session is a multi-turn dialogue in which the mentor passes the mentee various information.
In a session, the mentor can give instructions about relevant coding practices in Python that the mentee should follow when performing a task. For example, in Figure \ref{fig:memoryCode_fig1_sketch}, the instruction is the text in red on Day 1.
Once introduced, an instruction can be updated over time: in the case of Figure \ref{fig:memoryCode_fig1_sketch}, an update might be to  \textit{not} add the date anymore. 
When the mentee is asked to perform a task, it should remember and follow all the relevant instructions. 
Sessions can also include topics irrelevant to the target tasks: We refer to these topics as \textit{fillers} (in Figure \ref{fig:memoryCode_fig1_sketch}, the information in blue provided by the mentor).
Finally, a dialogue \textit{history} is the concatenation of all the sessions between the mentee and the mentor.

The dataset evaluates the models' ability to leverage the relevant instructions received throughout the history to perform the assigned tasks. 
To create dialogue histories we relied on both manual and automatic labor, thus optimizing quality and minimizing costs and effort, as described below. 

\begin{figure*}[t]
    \centering
    \includegraphics[width=0.9\linewidth]{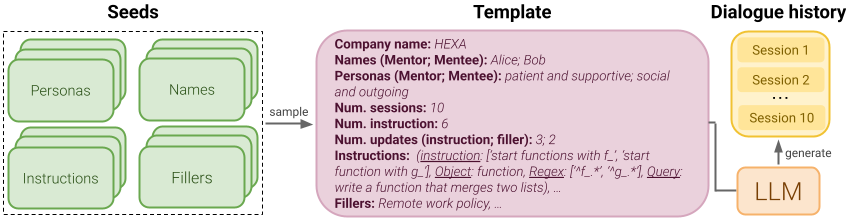}
    \caption{Dataset generation process. First, we randomly sample from our seeds to fill the variables of the template. The LLM is then prompted with this template to generate the dialogue history.}
    \label{fig:dataset_creation_process}
\end{figure*}

\subsection{Seeds}
\label{subsec:seeds}

A dialogue history is created based on a few crucial elements, or \textit{seeds}. We use four types of seeds:  
\textbf{instructions}, 
\textbf{fillers}, mentor and mentee \textbf{personas}, and \textbf{names}. 
For each seed, we define a set of possible values, from which we sample to generate histories (see Figure~\ref{fig:dataset_creation_process}). %
The possible seed values were manually defined by the authors to ensure high quality. Below, we describe each seed.

\paragraph{Instructions}
These are the coding instructions that the mentee must follow when generating a piece of code. 
An example is: \textit{always start function names with `g\_'.} Instructions are designed to be as simple as possible---recall that
we are interested in assessing the models' ability to leverage the information that emerged during interactions, not 
their ability to perform complex tasks. 

Each instruction applies to a specific Python object (e.g., \textit{function}).
Also, for some instructions, we define \textit{updates}:
For example, the instruction above would be updated to \textit{always start function names with `h\_'}.
Then, for each instruction,
we create an evaluation query and a test function.\footnote{Note that the evaluation query is the same for an instruction and for its updates.} Queries are specifically designed to trigger code generation that is relevant to the instruction (e.g., \textit{write a function that merges two sorted lists}).
Test functions are regular expressions: they only assess if the relevant instructions were followed (e.g., if the function starts with the required letter), and not the overall quality of the generated code.
We manually crafted 51 instructions, 16 of which can be updated up to 8 times, while the remaining ones do not have updates. We report the full list of instructions in Table \ref{tab:instructions_list} of Appendix \ref{app:seeds}.

\paragraph{Fillers} In real-world scenarios, interactions between colleagues can also include topics that do not necessarily impact daily tasks. To simulate this, we use fillers, which can be of two types. 
The first type contains general information about common topics at work such as \textit{remote work policy}.
The second contains instructions similar to those introduced above, but not strictly related to code generation, like \textit{use GitHub as the main version control system}. These latter fillers are meant to be harder distractors, as a model, recognizing them as instructions, might focus on them.

Fillers can be updated throughout sessions, however---unlike instructions---they are never evaluated.\footnote{For this reason, 
from now on,
`instructions' will always refer to 
coding ones---not fillers---unless
differently specified.}
We manually gathered 80 fillers, 50 of the first type, and 30 of the second. A filler can be updated up to 4 times. The full list of fillers are in Table \ref{tab:fillers} of Appendix \ref{app:seeds}.

\paragraph{Personas} Personas define the personality traits of the mentor and the mentee. 
By having different personas and combining them, we can generate conversations that are more diverse and thus increase the variety of the dataset. We define 6 personas for the mentor and 5 for the mentee (see Table \ref{tab:mentee_personas} and Table \ref{tab:mentor_personas} of Appendix \ref{app:seeds}). 

\paragraph{Names} We define lists of fictitious names for mentors, mentees, and companies, from which we randomly sample to generate the conversations (see Table \ref{tab:names} of Appendix \ref{app:seeds}). 

\subsection{Dialogue Histories}
\label{subsec:dataset_creation}

We generate the dialogue histories in two steps: we first create templates by sampling different combinations of seeds and other parameters, and then generate the actual histories based on these templates using an LLM, as shown in Figure \ref{fig:dataset_creation_process}.

\begin{table}[]
\begin{tabular}{ll}
\toprule
\textbf{Parameter}      & \textbf{Range}                                                                   \\
\midrule
Sessions ($n$)          & $$\{1,2,3,4,5,10,15,$$ \\ & $$20,30,40,50,100\}$$ \\
Sessions with instr. (\%)               &   $[50, 70]$                                                                    \\
Instr. in a session ($n$) &  $\{1,2,3\}$                                                                     \\
Instr. updates (\%)      & $[30, 70]$                                                                       \\
Filler updates (\%)      & $[50, 70]$     \\                          
\bottomrule
\end{tabular}
\caption{Parameters for dialogue history generation.}
\label{tab:parameters_range}
\end{table}

\paragraph{Template generation}  
We initially sample a \textbf{name} and a \textbf{persona} for the mentor and mentee, and a \textbf{name} for the company from our seeds. 
We then randomly pick a value for each of the following parameters: 
(i) \textbf{sessions}: how many sessions will be included in the dialogue history;
(ii) \textbf{sessions with instructions}: the percentage of sessions that will include an instruction. Since we set the maximum value to 70\%, some sessions will only have fillers;
(iii) \textbf{instructions in session}: how many instructions a session will include (min 1; max 3);
(iv) \textbf{instructions and update ratio}: the actual instructions that will be included, and how many of them will be updated;
(v) \textbf{fillers and update ratio}: same as for instructions.
Table \ref{tab:parameters_range} presents the parameters range we used to generate the dataset.

\begin{table}[]
    \centering
    \begin{tabularx}{1\columnwidth}{Xll}
    \toprule
    \textbf{Parameter}  & \textbf{Short dataset} & \textbf{Long dataset} \\
    & \textbf{(<15 sessions)} &  \textbf{(>15 sessions)} \\
    \midrule
    Sessions             & \normalsize{5.71} \scriptsize{($\pm$4.65)}            & \normalsize{48.00} \scriptsize{($\pm$27.85)}         \\
    Sessions\textsubscript{w/ instr.} & \normalsize{3.38} \scriptsize{($\pm$2.66)}            & \normalsize{28.13} \scriptsize{($\pm$16.56)}         \\
    Instr.               & \normalsize{4.98} \scriptsize{($\pm$4.10)}            & \normalsize{42.24} \scriptsize{($\pm$25.37)}         \\
    Instr.\textsubscript{added}           & \normalsize{3.56} \scriptsize{($\pm$2.62)}            & \normalsize{24.82} \scriptsize{($\pm$15.06)}         \\
    Instr.\textsubscript{updated}        & \normalsize{1.41} \scriptsize{($\pm$1.97)}            & \normalsize{17.42} \scriptsize{($\pm$11.93)}         \\
    Fillers              & \normalsize{5.04} \scriptsize{($\pm$4.75)}            & \normalsize{45.06} \scriptsize{($\pm$29.36)}         \\
    Filler\textsubscript{added}          & \normalsize{3.36} \scriptsize{($\pm$2.92)}            & \normalsize{24.63} \scriptsize{($\pm$12.70)}         \\
    Filler\textsubscript{updated}       & \normalsize{1.52} \scriptsize{($\pm$1.81)}            & \normalsize{18.86} \scriptsize{($\pm$13.48)}         \\
    Tokens         & \normalsize{3.20k} \scriptsize{($\pm$2.71k)}            & \normalsize{26.15k} \scriptsize{($\pm$15.50k)}         \\
    Vocabulary           & \normalsize{8.54k}                                    & \normalsize{14.24k}                                  \\
    \bottomrule
    \end{tabularx}
    \caption{Summary statistics (averages and standard deviations) for the `short' and `long' datasets.}
    \label{tab:dataset_statistics}
\end{table}

\paragraph{Dialogue history generation} For each session, we automatically construct a prompt incorporating the information from the template. The prompt introduces the company, the mentor, and the mentee, as well as the instructions and fillers of the session. We then use Command R+ \citep{cohere2024} to generate the session. We report examples of prompts in Table \ref{tab:prompt_conv_70} and \ref{tab:prompt_conv_108} of Appendix \ref{app:examples}. 

The resulting dataset contains 360 dialogue histories, 30 for each of the following number of sessions: 1, 2, 3, 4, 5, 10, 15, 20, 30, 40, 50, 100.
In what follows, we use `short' to refer to histories with fewer than 15 sessions (54\% of the total), and `long' to those with more than 15 sessions (46\%). Note that the longest history contains 63k tokens, which still fits the context window of all the models we used.
In Table \ref{tab:dataset_statistics}, we report the main statistics of the dataset. 
During the dataset creation, to ensure quality, we performed several generation rounds that we manually assessed and used to further optimize the prompting. Manual inspection of the final generated dialogue histories confirmed the overall quality and coherence of the dataset.

\section{Experiments}
\label{subsec:experiments}

We evaluate models on \textsc{MemoryCode} on three evaluation setups, each of them including a different kind of textual input.

\paragraph{\textsc{Instruction}} The input consists of a single instruction (e.g., in Figure~\ref{fig:memoryCode_fig1_sketch}, `\textit{add the current date to the start of every function}'). This setting is included to assess how good models are at performing coding tasks without any conversational setup.

\paragraph{\textsc{Session}} The input is an entire session (in Figure~\ref{fig:memoryCode_fig1_sketch}, a whole-day mentor-mentee interaction). 
In this setup, the model output is correct only if the model simultaneously adheres to \textit{all} the instructions introduced in the session.

\paragraph{\textsc{History}} The input of the model is the whole dialogue history, i.e., the concatenation of all sessions (in Figure~\ref{fig:memoryCode_fig1_sketch}, the entire 20-day mentor-mentee interaction). This setup is the most challenging one,
as it evaluates the ability to recall information from previous sessions and to use it together with new information to correctly perform the task. As such, it 
mimics realistic working scenarios, where colleagues interact over long periods.\\

Given an instruction and the model output, we assess it using the corresponding regex function. 
The model receives a score of 1 only if the instruction is correctly applied to all instances of the relevant Python object and there are no syntax errors.\footnote{Additionally, if the relevant Python object is not present in the generated code (<1\% of the cases), the instruction is not taken into account when averaging the scores.} For example, if the instruction is \textit{always start function names with `g\_'}, all functions in the generated code must start with `g\_'.
The overall model's performance is computed using macro-averaged accuracy.

\subsection{Models}

We test several recent LLMs on our benchmark, namely, three versions of Llama-3.1~\cite[8B-Instruct, 70B-Instruct, and 405B-Instruct;][]{dubey2024llama},
Command R+ \citep{cohere2024}, Command A \citep{cohere2025command}, GPT-4o~\citep{openai2024}, DeepSeek-V3 \citep{liu2024deepseek}, DeepSeek-R1 \citep{guo2025deepseek}. Our model selection includes both proprietary and open-weights models, as well as reasoning and non-reasoning models, and a broad range of model sizes. 
This provides us with a comprehensive overview of how various types of LLMs perform on our dataset. We note that all the models have been trained on code and tested on Python coding benchmarks such as HumanEval \citep{chen2021evaluating} and MBPP \citep{austin2021program}. The details to reproduce the results are provided in Appendix~\ref{app:hyperparameters}.

\section{Results}
\label{sec:experiments}

\begin{figure*}[tb]
    \centering
    \includegraphics[width=\linewidth]{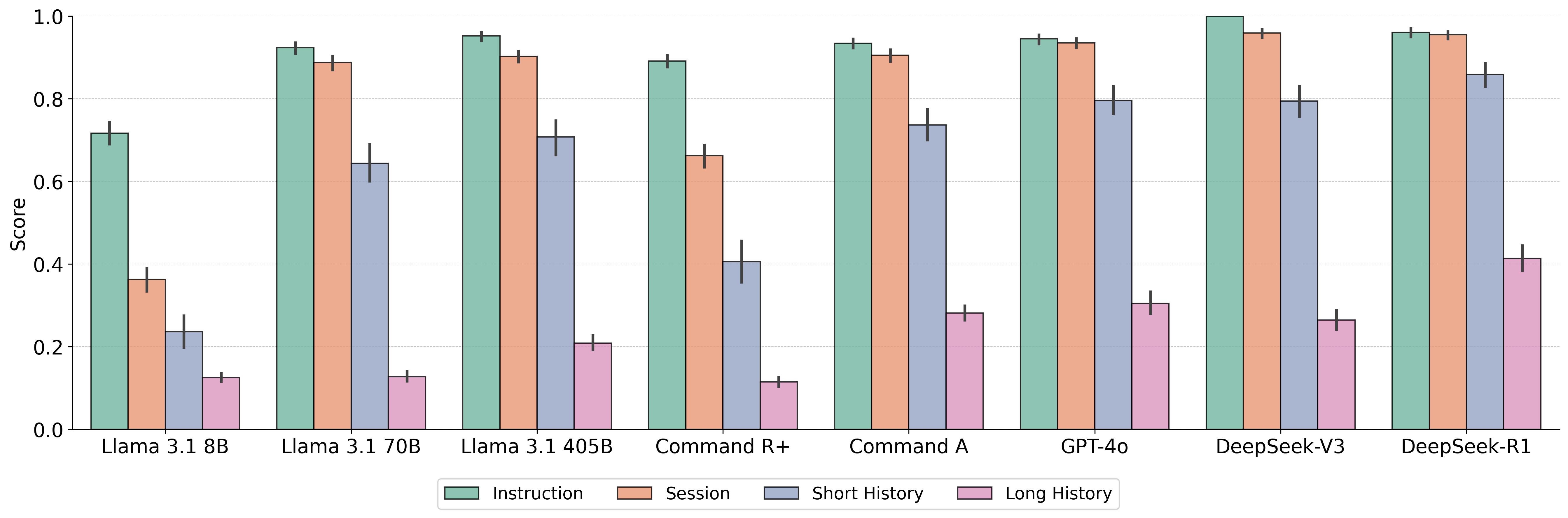}
    \caption{Average \textsc{Instruction}, \textsc{Session}, and \textsc{History} scores per model. For the latter, `short' includes dialogue histories with less than 15 sessions, `long'  those with 16 to 100 sessions. Results include 95\% confidence intervals.}
    \label{fig:average_score}
\end{figure*}

In this section, we report the performance of the models across the evaluation setups described in Section~\ref{subsec:experiments}. Figure~\ref{fig:average_score} shows the average score for each model. The exact numbers are included in Table \ref{tab:detailed_scores} of Appendix \ref{sec:detailed_scores}. 

\paragraph{\textsc{Instruction}} As shown in Figure~\ref{fig:average_score}, all models achieve high or very high accuracy on this setup. 
This aligns with our goals of having relatively easy instructions.
In particular, all the large models approach or exceed 0.9 accuracy score---reported in the $[0,1]$ scale--- with DeepSeek-V3 achieving a perfect. 
While the results of Llama-8B are lower, they still show that even a small, non-specialized model achieves good performance, confirming the easiness of the task.

As a sanity check, we run a set of experiments in which we do not prompt the models with the necessary instructions (e.g., \textit{use CamelCase}), but directly run the evaluation (in this case, we check if CamelCase was used).\footnote{
Due to budget limits, we only used Command R+ for this experiment and for the analysis in Section \ref{subsec:retrieval_or_reasoning_problem}.
We expect the results to be representative of all other models' behavior.} This setup verifies that models do not solve \textsc{MemoryCode} through their default behavior. Models fail spectacularly, achieving an extremely low average accuracy (consistently lower than 0.01), confirming that the instructions we provide are crucial to executing the tasks correctly. 

\paragraph{\textsc{Session}} The performance in this setup is very similar to \textsc{Instruction} for the larger Llama models, DeepSeek models, Command A, and GPT-4o, indicating that these models have no difficulties at retrieving the relevant information in a single session. Command R+ shows a larger drop of 0.22 (25\% relative drop compared to \textsc{Instruction}), while Llama-3.1-8B shows a major drop of 0.34 (48\%), which indicates its inability to retrieve relevant information across multiple turns.

\paragraph{\textsc{History}} Things change dramatically in this setup, with a degradation in performance across the board already for `short' dialogue histories. In particular, GPT-4o shows a drop of 0.14. A similar decrease is observed in the larger LLaMA models, DeepSeek models, and Command R. The drop in score is even greater for the other models: 0.47 (67\% relative drop compared to \textsc{Instruction}) for Llama-3.1-8B and 0.48 (54\%) for Command R+. 
These results indicate that as the number of sessions increases, even the best-in-class models have difficulties in identifying and applying the relevant instructions. 

A more dramatic drop is observed in the `long' setup. Here, all the models struggle, with GPT-4o, the best non-reasoning model, only achieving 0.30 accuracy,
which indicates a relative drops of 61\% from 
the `short' setup and of 
67\% 
from \textsc{Instruction}. While DeepSeek-R1 significantly improves over the performance of GPT-4o, it still shows a drop of 0.56 points compared to the performance on the \textsc{Instruction} setup. These results confirm that the main challenge of the benchmark is not merely following individual instructions but reasoning over a sequence of instructions. Moreover, they highlight that even newer models equipped with reasoning abilities struggle on this benchmark, suggesting that complementary or alternative capabilities are needed to solve it effectively.
The drop is even more significant for the other models: the performance of Llama-3.1-405B drops by 78\% compared to \textsc{Instruction}, Command R+ by 87\%. 
Crucially, this happens even though the tasks on which models are evaluated are identical to those in the \textsc{Instruction} and \textsc{Session} setups, where the models achieved nearly-perfect accuracy. 
The difference in performance, hence, is to be ascribed to models' inability to retrieve and reason over 
relevant pieces of information present in their input.

\section{Analysis}
\label{sec:analysis}

In this section, we conduct an analysis aimed at understanding which factors influence model performance on the \textsc{MemoryCode} benchmark. We focus on the \textsc{History} setup, the most important and challenging one. 

\begin{figure}[hbt!]
    \centering
    \includegraphics[width=\linewidth]{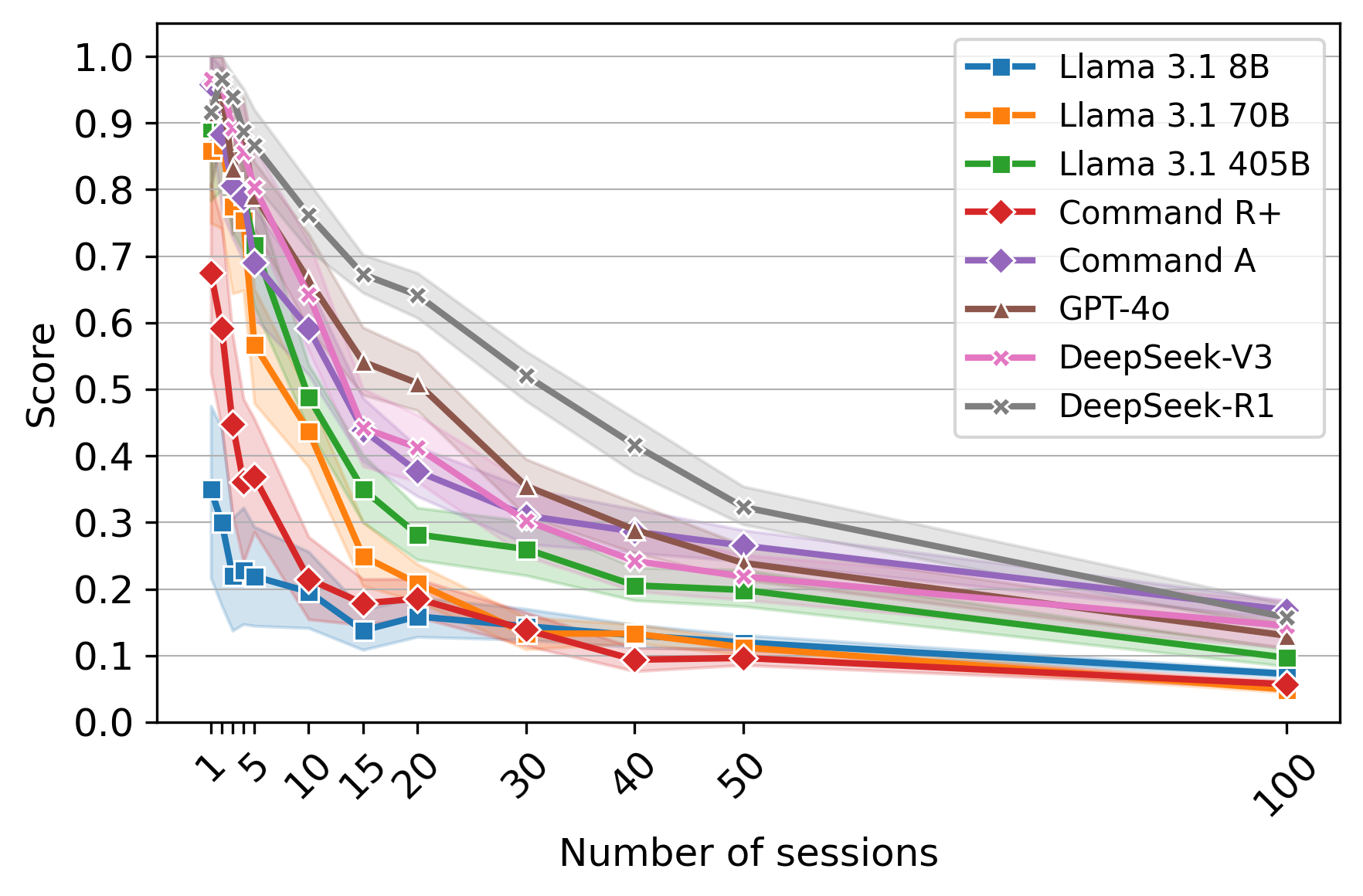}
        \caption{Score per number of sessions.}
    \label{fig:per_session_score_all_models}
\end{figure}

\subsection{Effect of Number of Sessions} 
\label{subsec:effect_number_sessions}

In Figure \ref{fig:per_session_score_all_models}, we show how the performance decreases with an increasing number of sessions. 
Consistently with the aggregated patterns shown in Figure~\ref{fig:average_score}, 
relevant variations can be observed across models when the number of sessions is rather low, which reflects the differences observed in the `short' setup. 
However, all models converge to a similarly, extremely low accuracy (around 0.1) when the number of sessions approaches 100. This confirms that all models are similarly bad at handling requests involving long-context inputs. 
 
This weakness could be due to limitations in retrieving the relevant information from the dialogue history, compositionally applying instructions from the chain, inferring their latest updates, or any combination of these factors. 
Below, we shed light on this issue. 

\subsection{Retrieval or Multi-Instruction-following Problem?}

\label{subsec:retrieval_or_instruction_following_problem}

\begin{figure}[htb]
    \centering
    \includegraphics[width=\linewidth]{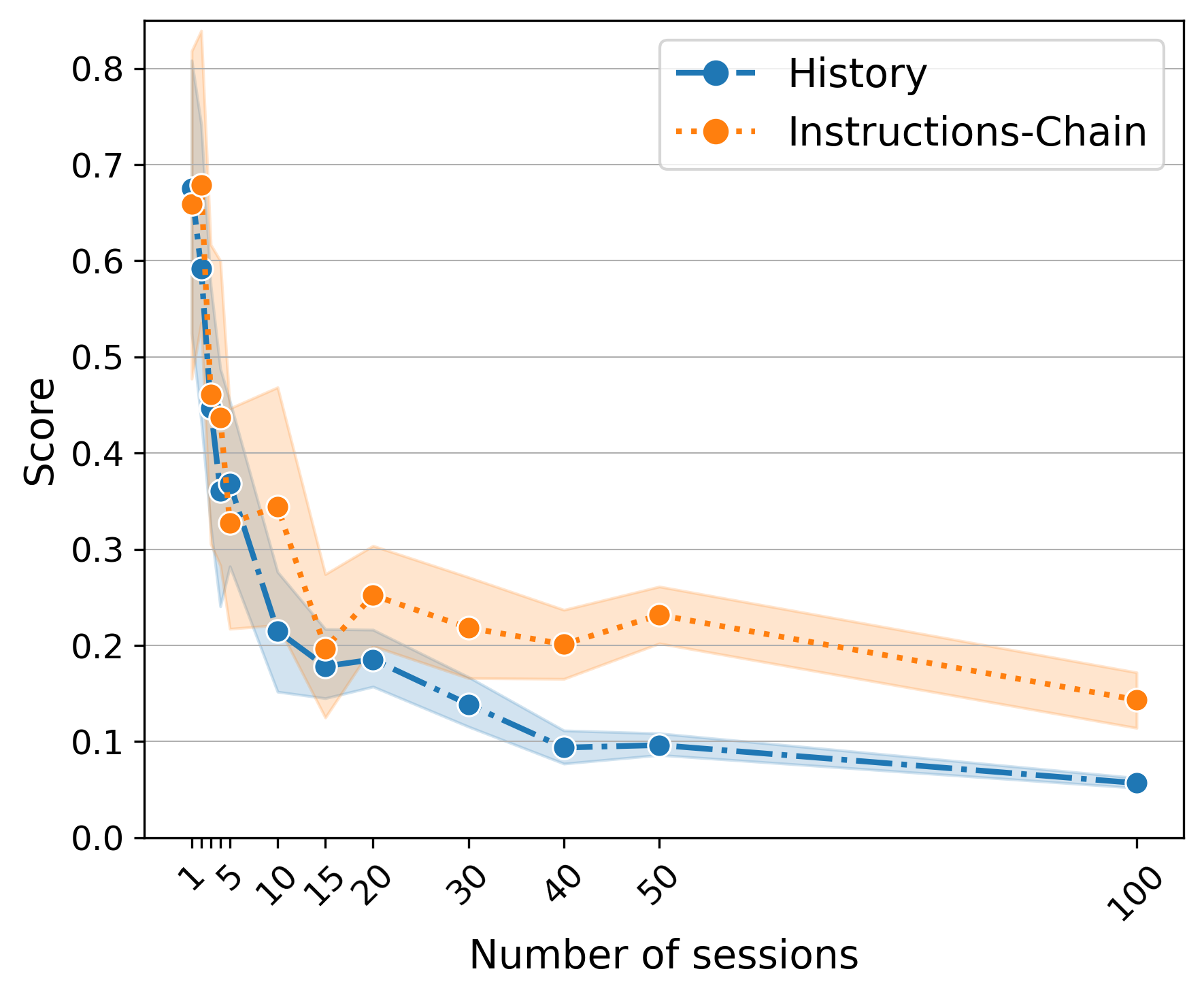}
    \caption{Per-sessions score for \textsc{Instr.-chain}.}
    \label{fig:per_session_command_r_plus}
\end{figure}

If the poor performance in the \textsc{History} setup was due to retrieval, then passing models only the full chain of instructions and updates---without any intervening irrelevant text---should solve the issue. 
\textit{Vice versa}, 
if the issue stemmed from difficulties in understanding and jointly applying multiple instructions,
they should still perform poorly.
We test these assumptions by feeding Command R+ with only the entire chain of instructions needed to solve a task. We name this setup \textsc{Instructions-chain}. 
As shown in Figure~\ref{fig:per_session_command_r_plus}, the trend is strikingly similar to the one observed in \textsc{History}, with the model still struggling even if only the relevant information is provided, with no dialogue history.
This indicates that models' drop in performance is mainly due to their inability to reason compositionally over a sequence of instructions. 

Retrieval from the dialogue history also plays a role, as indicated by the slightly higher performance in \textsc{Instructions-chain} over \textsc{History}. To mitigate this retrieval issue, we experimented with Retrieval Augmented Generation (RAG), where instead of providing as input the whole history, we retrieve its relevant parts only and feed them to the model. However, we did not observe any improvement over \textsc{History} (see Appendix \ref{app:rag_experiments}).

\subsection{Effect of Instruction Updates}

After delineating the contribution of retrieval and multiple-instruction-following to the performance degradation, we now examine the role of instruction updates.

We define the \textbf{update rank} of an instruction as the number of times the instruction is updated throughout the dialogue history, for both `short' and `long' setups. An update rank of 0 means that the instruction was never updated. The maximum number of updates in our dataset is 7.

The number of updates increases with number of sessions: to control for the latter, we compute the scores for the different update ranks for each number of session independently. 

\begin{figure}
    \centering
    \includegraphics[width=\linewidth]{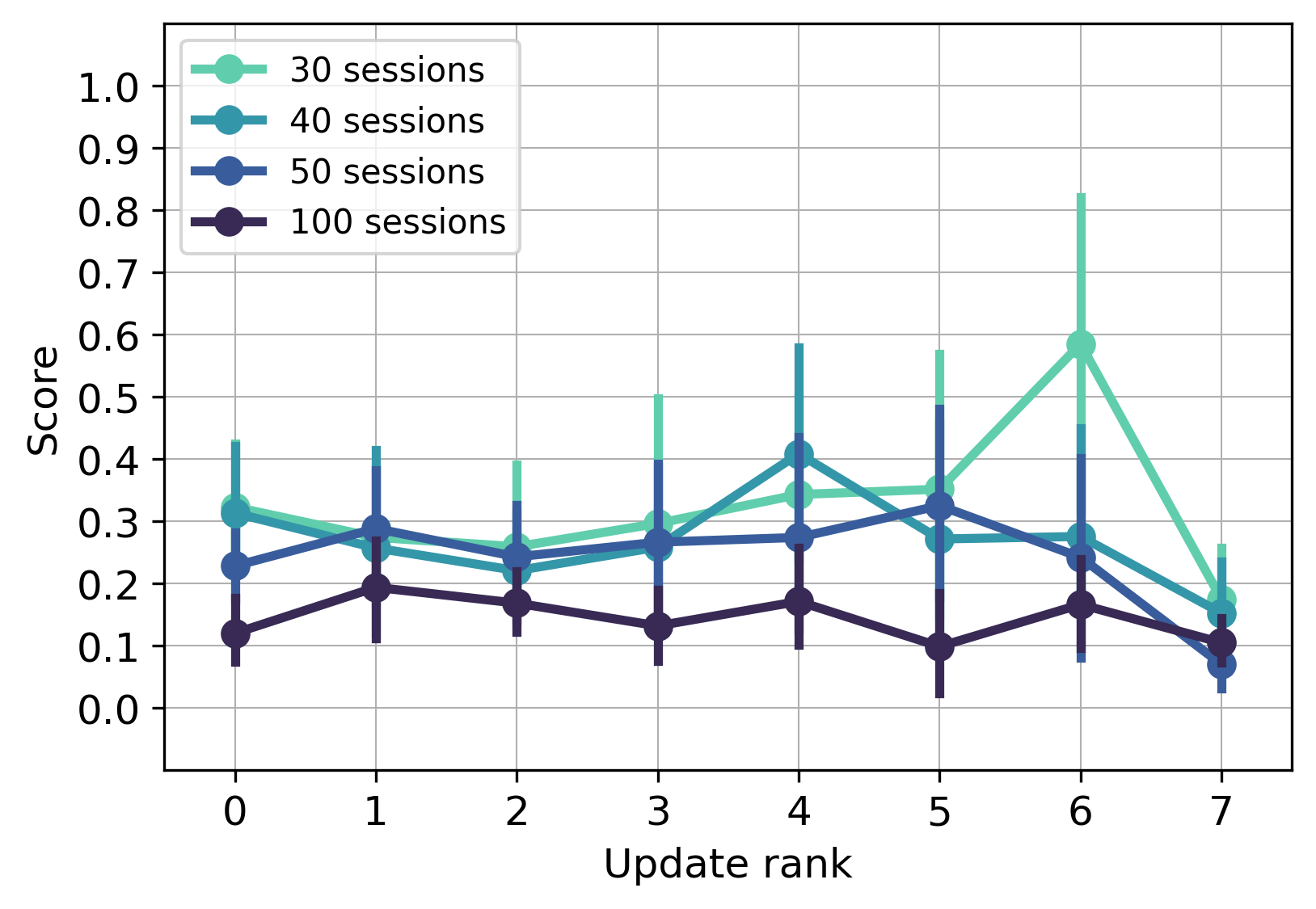}
    \caption{Score as a function of update rank.}
    \label{fig:update_rank}
\end{figure}

Figure \ref{fig:update_rank} reports the result for each update rank obtained with GPT-4o. 
The results show that the score is stable across different update ranks. 

This suggests that the poor performance on this dataset is primarily due to the model's difficulty in jointly following multiple rules, rather than its ability to update a single instruction.

\subsection{Instruction Difficulty}
We finally assess if, besides the updates,
the very nature of each instruction makes it more challenging than others.
For this, we compute the average per-instruction score of the best-performing model, GPT-4o, for both the initial instruction and its updates.  
The results reveal notable variations across instructions: 
In particular, instructions and updates that are less common in practice---like those including a digit in object names---tend to obtain lower scores compared to more common ones, such as using docstrings or annotations. 
Figure~\ref{fig:per_pivot_score_insertion} and~\ref{fig:per_pivot_score_update} in Appendix~\ref{app:per_instruction_scores} report the results of this analysis. %
Overall, this suggests that the models' ability to retrieve and update relevant instructions is modulated by their characteristics and that familiarity is likely to play a role. 
An example (from GPT-4o) is shown in Table~\ref{tab:qual_example_2}. 
In this case, the model correctly follows the instruction on comments, but not those about attribute and variable names.

We note that instruction difficulty naturally interacts with another key aspect of long-term interactions: the degree of compositionality among the pieces of information provided by the mentor. Besides overwriting prior instructions or appending complementary ones, mentors may issue instructions that combine in compositional ways. For instance, one instruction might require that any line of code using parallel computing libraries (e.g., NumPy or Torch) be followed by a comment, while another might call for merging comments from adjacent lines if the associated code is highly correlated. Although we do not explicitly analyze such cases here, we argue that our approach can be readily adapted to systematically study these scenarios in a controlled setting.

\arrayrulecolor{gray!50}
\begin{table}[t!]
\centering
\begin{tabularx}{1\columnwidth}{X}
\toprule
\textbf{DIALOGUE HISTORY} \\
\midrule
(1) \textbf{Yoon-Seo:} Alexander, it's great to finally meet you. I'm Yoon-Seo, your mentor  \dots \\
\hdashline
(2) \textbf{Yoon-Seo:} \dots One of our key practices is to \colorbox{Apricot}{always add comments to your code}\dots \\
\textbf{Yoon-Seo:} \dots Another convention we follow is to \colorbox{CornflowerBlue}{end variable names with `\_n'} \dots \\
\hdashline
(3) \textbf{Yoon-Seo:} \dots \colorbox{CornflowerBlue}{All variable names should} \colorbox{CornflowerBlue}{now end with `\_z'} \dots \\
\hdashline
(4) \textbf{Yoon-Seo:} \dots When \colorbox{YellowGreen}{naming attributes}, [...] \colorbox{YellowGreen}{include the string `chx'} in the name \dots \\
\midrule
\textbf{EVALUATION QUERY} \\
\midrule
Write a function that checks if a string is a valid palindrome \\
\midrule
\textbf{MODEL OUTPUT} \\
\midrule
\texttt{def is\_palindrome\_\colorbox{YellowGreen}{chx}(s\_\colorbox{YellowGreen}{chx}):} \\    
\texttt{    \colorbox{Apricot}{\# Initialize two pointers}} \\   
\texttt{    left\_\colorbox{YellowGreen}{chx}\colorbox{CornflowerBlue}{\_n} = 0} \\    
\texttt{    right\_\colorbox{YellowGreen}{chx}\colorbox{CornflowerBlue}{\_n} = len(s\_\colorbox{YellowGreen}{chx}) - 1} \\
\texttt{    \dots} \\
\bottomrule
\end{tabularx}
\caption{A 4-session dialogue history (session id in parenthesis). 
GPT-4o correctly applies the instruction about \colorbox{Apricot}{comments} but not the one about \colorbox{YellowGreen}{attribute names} and the update on \colorbox{CornflowerBlue}{variable names}.}
\label{tab:qual_example_2}
\end{table}

\section{Conclusions}
\label{sec:discussion}

In this paper, we proposed \textbf{\textsc{MemoryCode}}, 
a new benchmark to assess state-of-the-art LLMs in their ability to retrieve and reason over pieces of information in multi-session dialogue histories reflecting real-world scenarios.
Differently from many existing datasets, the tasks in \textsc{MemoryCode} do not require any complex reasoning, and are easily solved by the models when provided in isolation. 
The main challenge of \textsc{MemoryCode} lies in the ability to keep track of multiple simple instructions received throughout a multi-session interaction, and to jointly apply them to perform a task.
When the number of sessions is small (<15), SotA models like GPT-4o manage to perform the task well. However, as the number grows up to 100, even these models face a dramatic drop in performance. 
Our analysis shows that this is mainly due to their failure to reason over a long chain of simple instructions. 

Overall, our results show a severe limitation of current LLMs. 
The inability to keep track of simple information as the interaction with a human unfolds effectively hinders their adoption in real-world scenarios and restricts their usage to addressing 
single, self-contained problems. 
We argue that effective long-term collaboration cannot be achieved by further scaling alone, be it model or dataset size, context window, or test-time compute.
Rather, we believe that new mechanisms to handle and retrieve from long-term memory need to be developed. 
\textsc{MemoryCode} contributes to this challenging and yet crucial goal, by providing a robust benchmark for developing and testing such methods.

\section*{Limitations}

While \textsc{MemoryCode} and our experimental setup enable us to identify key strengths and weaknesses of current models, certain limitations remain, which could inspire future research. 
First, \textsc{MemoryCode} is based on synthetic data. 
This choice was driven by both cost considerations---as collecting real interactions would have been much more expensive---, and the need for greater control over the factors influencing model performance. 
However, future work could explore more realistic interactions by relaxing the constraints imposed in \textsc{MemoryCode}. 
Second, our experiments do not establish a human performance upper bound. This could be an interesting direction for future investigation for future work, as it would provide useful information on human limitations at keeping track of relevant information provided over long periods of time and amidst large amount of irrelevant information.
Third, our dataset only focus on a specific kind of task, namely, coding. This kind of task was chosen due to it being very common in real-world scenarios and very easy to evaluate. However, future work should expand to other domains, to assess if results are consistent with those that we report. 
While we are aware of the limitations above, and that others possibly exist, we believe these do not impact the robustness of our findings.

\section*{Acknowledgments}
UPF was funded by the European Research Council (ERC) under the European Union’s Horizon 2020 research and innovation programme (grant agreement No. 101019291). During his work at the University of Amsterdam, Alberto Testoni was supported by funding from the European Research Council (ERC) under the European Union’s Horizon 2020 research and innovation programme (grant agreement No. 819455, PI R. Fernández). Alberto Testoni is currently funded by the project CaRe-NLP (PI Iacer Calixto) with file number NGF.1607.22.014 of the research programme AiNed Fellowship Grants, which is (partly) financed by the Dutch Research Council (NWO).This paper reflects the authors’ view only, and the funding agency is not responsible for any use that may be made of the information it contains.

\bibliography{refs}

\newpage
\appendix
\arrayrulecolor{black}          
\begin{table*}
    \centering
    \begin{tabular}{lllll}
        \toprule
         \textbf{Model} &  \textbf{Instruction} &  \textbf{Session} & \textbf{Short History} & \textbf{Long History} \\
         \midrule
         
         Llama 3.1 8B &  71.7 \scriptsize{$\pm$1.35 }&  36.3 \scriptsize{$\pm$1.36 }&  23.6 \scriptsize{$\pm$1.94 }&  12.5 \scriptsize{$\pm$0.50 }\\
         
         Llama 3.1 70B &  92.3 \scriptsize{$\pm$0.66 }&  88.8 \scriptsize{$\pm$0.85 }&  64.4 \scriptsize{$\pm$2.34 }&  12.7 \scriptsize{$\pm$0.60 }\\
         
         Llama 3.1 405B &  95.2 \scriptsize{$\pm$0.51 }&  90.2 \scriptsize{$\pm$0.65 }&  70.8 \scriptsize{$\pm$2.02 }&  20.9 \scriptsize{$\pm$0.88 }\\

          Command R+ &  89.1 \scriptsize{$\pm$0.72 }&  66.3 \scriptsize{$\pm$1.29 }&  40.5 \scriptsize{$\pm$2.47 }&  11.4 \scriptsize{$\pm$0.58 }\\

          Command A &  93.4 \scriptsize{$\pm$0.61 }&  90.5 \scriptsize{$\pm$0.73 }&  73.7 \scriptsize{$\pm$1.89 }&  28.1 \scriptsize{$\pm$0.91 }\\

          GPT-4o &  94.5 \scriptsize{$\pm$0.57 }&  93.5 \scriptsize{$\pm$0.56 }&  79.6 \scriptsize{$\pm$1.67 }&  30.5 \scriptsize{$\pm$1.31 }\\

          DeepSeek-V3 &  \textbf{100.0 \scriptsize{$\pm$0.00}}&  \textbf{95.9 \scriptsize{$\pm$0.49 }}&  79.4 \scriptsize{$\pm$1.79 }&  26.4 \scriptsize{$\pm$1.26 }\\

          DeepSeek-R1 & 96.1 \scriptsize{$\pm$0.53 }&  95.5 \scriptsize{$\pm$0.47 }&  \textbf{85.9 \scriptsize{$\pm$1.35 }}&  \textbf{41.3 \scriptsize{$\pm$1.56 }}\\
         
         \bottomrule
    \end{tabular}
    \caption{Average \textsc{Instruction}, \textsc{Session}, and \textsc{History} percentage scores per model with standard errors. For the latter, `short' includes dialogue histories with less than 15 sessions, `long'  those with 16 to 100 sessions.}
    \label{tab:detailed_scores}
\end{table*}

\begin{figure}[t!]
    \centering
    \includegraphics[width=\linewidth]{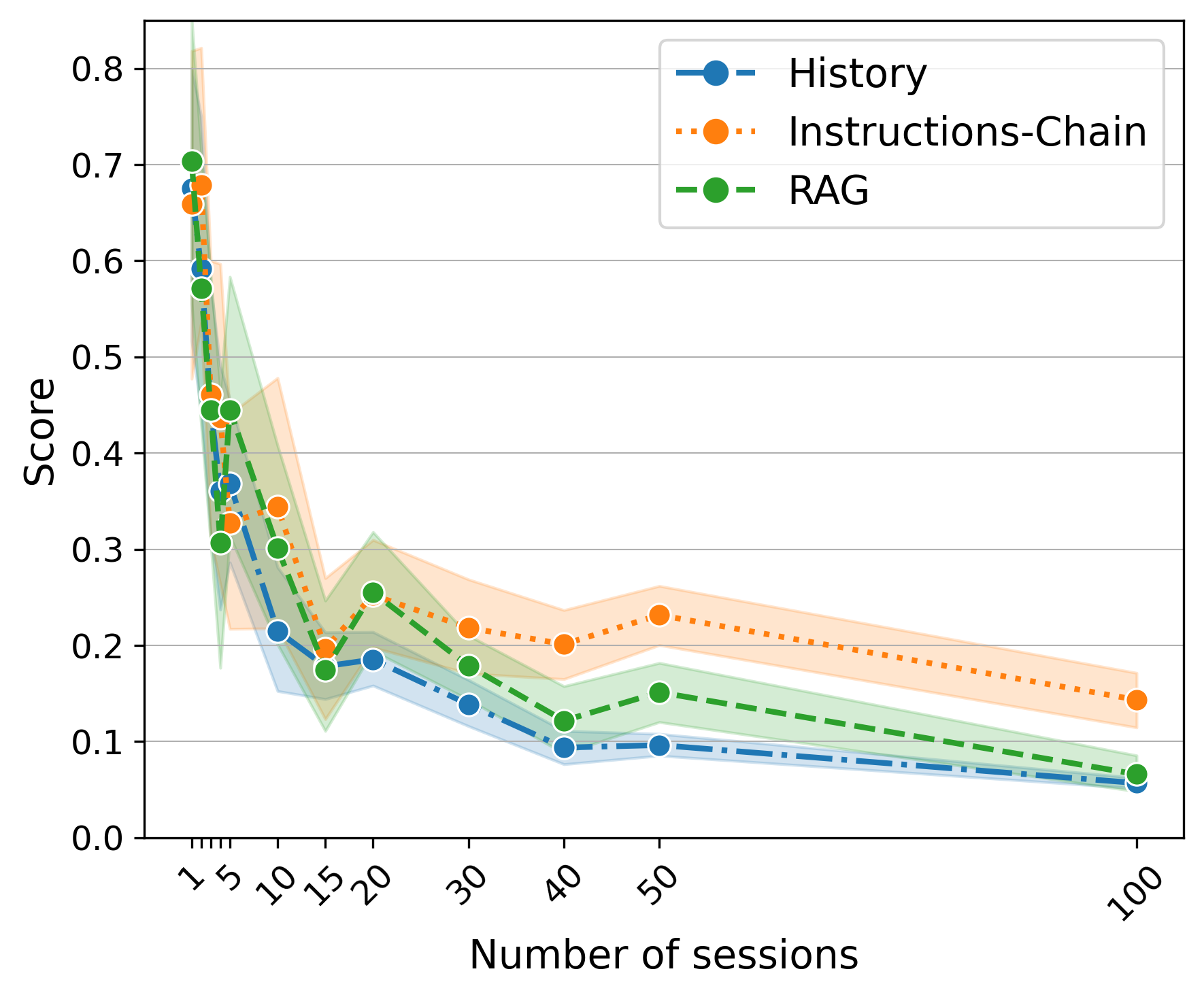}
    \caption{Score as a function of number of sessions. 
    }
    \label{fig:cum_rag_pivot_only}
\end{figure}

\section{Seeds}
\label{app:seeds}

\paragraph{Instructions} We used 51 instructions during data generation, as shown in Table \ref{tab:instructions_list}. Each instruction is applied to a specific Python object, and for pivots that can be updated, the update values are provided in brackets.

\paragraph{Fillers} The list of all 80 fillers used during data generation is provided in Table \ref{tab:fillers}. Some fillers can be updated over time.

\paragraph{Names} The list of mentor, mentee and company names used during data generation is provided in Table \ref{tab:names}.

\paragraph{Personas} The lists of mentor and mentee personas are provided in Table \ref{tab:mentor_personas} and Table \ref{tab:mentee_personas}.

\section{Detailed scores}
\label{sec:detailed_scores}
We provide the detailed scores of the different models evaluated on Instruction, Session,  Short History and Long History in Table \ref{tab:detailed_scores}. 

\section{Hyperparameters}
\label{app:hyperparameters}
We use Command R+ with a temperature of 0.9 and a top-p value of 0.9 to generate the conversations. During evaluation, we use a temperature of 0 for all models. 

\section{RAG experiments}
\label{app:rag_experiments}
We performed the Retrieval-augmented Generation (RAG) experiments using the rerank-english-v3.0 model \citep{coherererank2024}. The basic unit for retrieval were the previous sessions. We also tried with different retrieval units, such as paragraphs and turns, and obtained similar results. 
We provide the model the top-\textit{k} retrieved sessions, where \textit{k} was defined as the number of sessions with pivots. Note that, by dynamically defining \textit{k} in this way, rathr than using a fix value, we facilitate the retrieval of the relevant sessions only, eliminating potential noise.
We report the results of RAG in Figure \ref{fig:cum_rag_pivot_only}, together with those of Instructions-only and Cumulative. RAG provides a marginal improvement over cumulative for `short' conversations, but it then converges with Cumulative for the `long' ones.

\section{Per-Instruction Scores}
\label{app:per_instruction_scores}
\begin{figure*}
    \centering
    \includegraphics[width=0.9\linewidth]{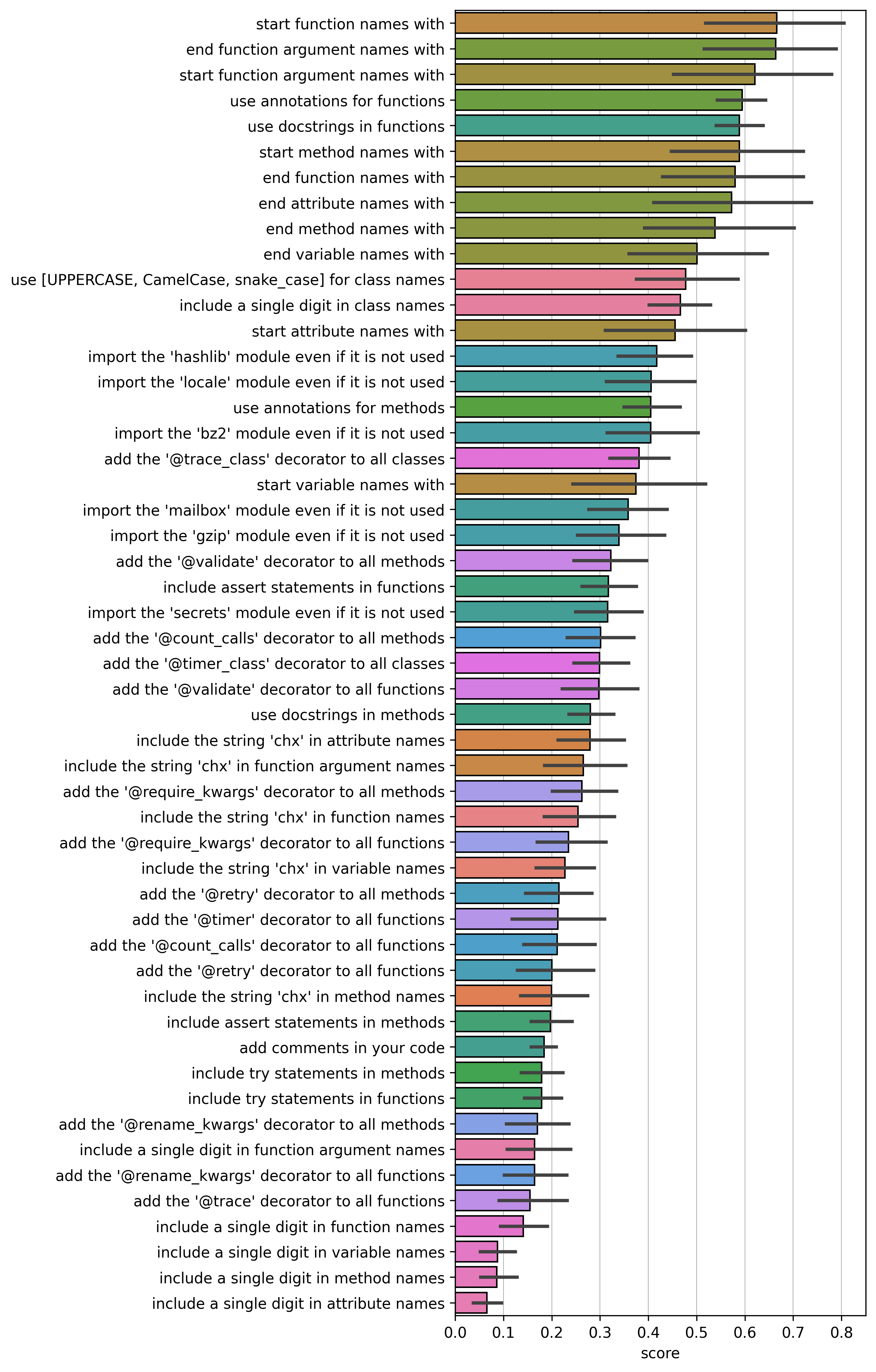}
    \caption{Average per-instruction insertion scores for GPT-4o. Results include 95\% confidence intervals.}
    \label{fig:per_pivot_score_insertion}
\end{figure*}

\begin{figure*}[htb!]
    \centering
    \includegraphics[width=\linewidth]{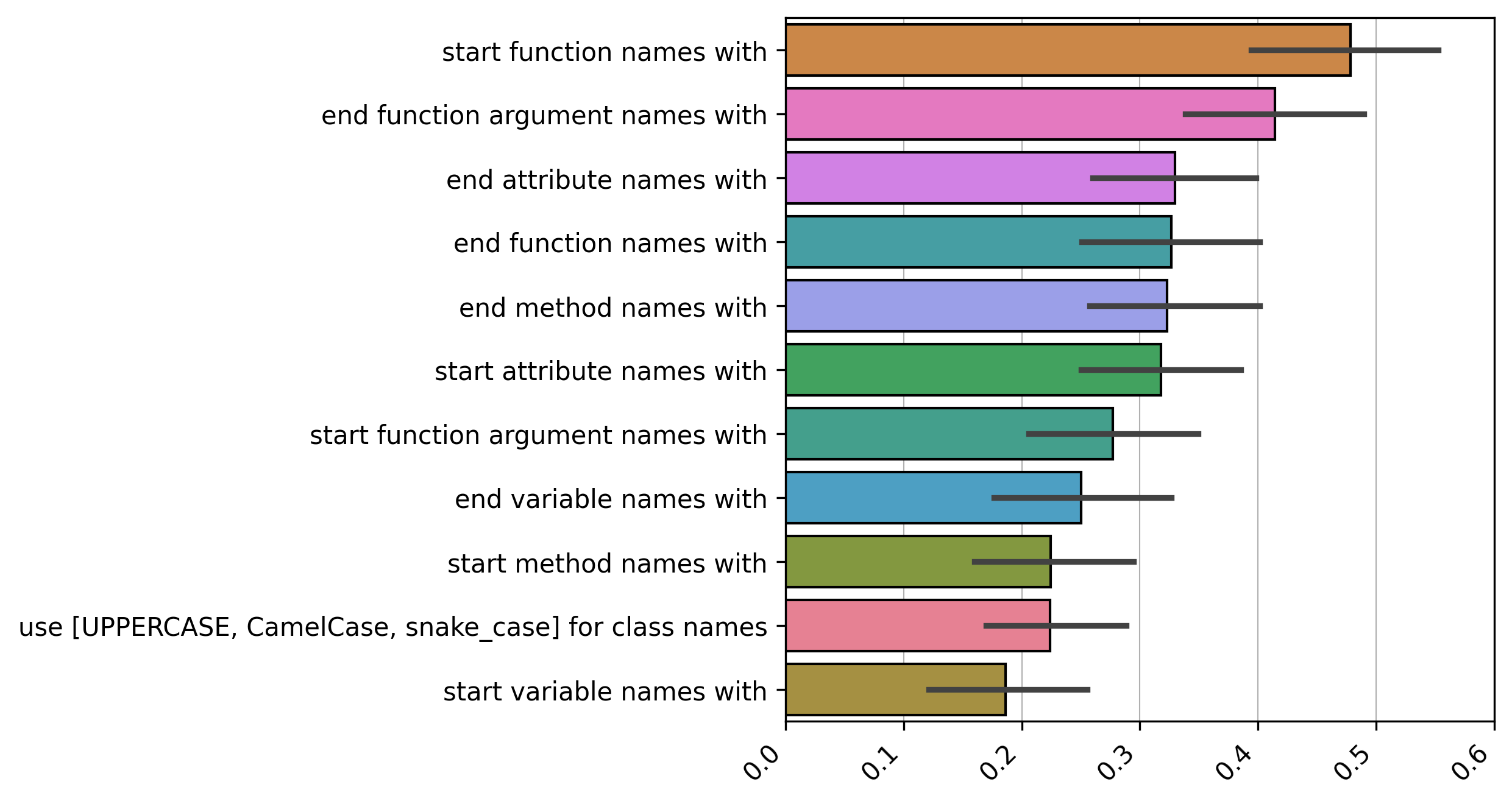}
    \caption{Average per-instruction update scores for GPT-4o. Results include 95\% confidence intervals.}
    \label{fig:per_pivot_score_update}
\end{figure*}

We report in Figure \ref{fig:per_pivot_score_insertion} the average results for each instruction, and in 
Figure \ref{fig:per_pivot_score_update} those for each update. Remember that only some of the instructions have an update (See Section \ref{subsec:seeds}).

\section{Examples}
\label{app:examples}
We provide examples of dialogue histories along with their corresponding prompts. The first one, shown in Table \ref{tab:conv_70}, is a 3-session history with 3 pivots, and its prompt is provided in Table \ref{tab:prompt_conv_70}. The second one, presented in Table \ref{tab:conv_108}, consists of 4 sessions with 3 pivots, with its prompt shown in Table \ref{tab:prompt_conv_108}. The system prompt and the session-level prompts were designed to ensure consistency across sessions. Additionally, we provide examples of prompts to get the model output for the Instructionm History and Instructions-Chains settings in Tables \ref{tab:intruction_prompt_example}, \ref{tab:conversation_prompt_example} and \ref{tab:pivot_only}.

\section{Computational budget}
\label{sec:computational_budget}
The prompts used to generate the dataset consist of a total of 0.9M tokens, while the dataset itself contains 4.6M tokens. The total cost of generating the dataset using Command R+ through Cohere's API is approximately \$50. The cost of evaluating a single model on the Instruction, Session and Cumulative settings using online APIs is approximately \$50.

\section{Ethics statement}
\label{sec:ethical statement}
The dataset we are releasing is synthetic and, therefore, does not contain any personally identifiable information. Moreover, we did not recruit any human participants, as we manually validated the quality of the dataset ourselves.

\onecolumn
\arrayrulecolor{gray!50}
\begin{xltabular}{\linewidth}{Xl}
    \specialrule{1.2pt}{0pt}{0pt}
    \rowcolor{gray!20} \textbf{Instruction} & \textbf{Object} \\
    \specialrule{1.2pt}{0pt}{0pt}
    use \{all UPPERCASE, CamelCase, snake\_case\} for class names & class \\
    \specialrule{0.6pt}{0pt}{0pt}
    include the string 'chx' in function names & function \\
    start function names with \{'a\_', 'b\_', 'c\_', 'd\_', 'x\_', 'y\_', 'fn\_', 'gn\_'\} & function \\
    end function names with \{'\_a', '\_b', '\_c', '\_d', '\_x', '\_y', '\_fn', '\_gn'\} & function \\
    \specialrule{0.6pt}{0pt}{0pt}
    include the string 'chx' in variable names & variable \\
    start variable names with \{'x\_', 'y\_', 'z\_', 'w\_', 'm\_', 'n\_', 'vr\_', 'wr\_'\} & variable \\
    end variable names with \{'\_x', '\_y', '\_z', '\_w', '\_m', '\_n', '\_vr', '\_wr'\} & variable \\
    \specialrule{0.6pt}{0pt}{0pt}
    start method names with \{'x\_', 'n\_', 'o\_', 'p\_', 'a\_', 'b\_', 'md\_', 'ud\_'\} & method \\
    end method names with \{'\_x', '\_n', '\_o', '\_p', '\_a', '\_b', '\_md', '\_ud'\} & method \\
    include the string 'chx' in method names & method \\
    \specialrule{0.6pt}{0pt}{0pt}
    include the string 'chx' in attribute names & attribute \\
    start attribute names with \{'q\_', 'r\_', 's\_', 't\_', 'i\_', 'j\_', 'at\_', 'xt\_'\} & attribute \\
    end attribute names with \{'\_q', '\_r', '\_s', '\_t', '\_i', '\_j', '\_at', '\_xt'\} & attribute \\
    \specialrule{0.6pt}{0pt}{0pt}
    start function argument names with \{'e\_', 'f\_', 'g\_', 'h\_', 'i\_', 'j\_', 'x\_', 'a\_'\} & function argument \\
    end function argument names with \{'\_e', '\_f', '\_g', '\_h', '\_i', '\_j', '\_x', '\_a'\} & function argument \\
    include the string 'chx' in function argument names & function argument \\
    \specialrule{0.6pt}{0pt}{0pt}
    use annotations for functions & function annotation \\
    include try statements in functions & function try \\
    include assert statements in functions & function assert \\
    use docstrings in functions & function docstring \\
    \specialrule{0.6pt}{0pt}{0pt}
    add the '@retry' decorator from the 'pedantic' module to all functions & function decorator \\
    add the '@count\_calls' decorator from the 'pedantic' module to all functions & function decorator \\
    add the '@rename\_kwargs' decorator from the 'pedantic' module to all functions & function decorator \\
    add the '@require\_kwargs' decorator from the 'pedantic' module to all functions & function decorator \\
    add the '@timer' decorator from the 'pedantic' module to all functions & function decorator \\
    \specialrule{0.6pt}{0pt}{0pt}
    use annotations for methods & method annotation \\
    include try statements in methods & method try \\
    include assert statements in methods & method assert \\
    use docstrings in methods & method docstring \\
    \specialrule{0.6pt}{0pt}{0pt}
    add the '@retry' decorator from the 'pedantic' module to all methods & method decorator \\
    add the '@count\_calls' decorator from the 'pedantic' module to all methods & method decorator \\
    add the '@rename\_kwargs' decorator from the 'pedantic' module to all methods & method decorator \\
    add the '@require\_kwargs' decorator from the 'pedantic' module to all methods & method decorator \\
    \specialrule{0.6pt}{0pt}{0pt}
    add comments in your code & comment \\
    \specialrule{0.6pt}{0pt}{0pt}
    import the 'secrets' module even if it is not used & import \\
    import the 'locale' module even if it is not used & import \\
    import the 'mailbox' module even if it is not used & import \\
    import the 'bz2' module even if it is not used & import \\
    import the 'gzip' module even if it is not used & import \\
    import the 'hashlib' module even if it is not used & import \\
    \specialrule{1.2pt}{0pt}{0pt} 
    
\caption{List of all the instructions used in pivots.}
\label{tab:instructions_list}
\end{xltabular}
\twocolumn

\onecolumn
\begin{xltabular}{\linewidth}{X}
    \specialrule{1.2pt}{0pt}{0pt}
    \rowcolor{gray!20} \textbf{Filler}  \\
    \specialrule{1.2pt}{0pt}{0pt}
Contract renewal negotiations and finalization \\ 
 \midrule
Planning engaging social activities for the upcoming holidays \\ 
 \midrule
Flexible work hours and their impact on work-life balance \\ 
 \midrule
Troubleshooting internet connectivity issues across different locations \\ 
 \midrule
Promoting a harmonious workplace through effective conflict resolution \\ 
 \midrule
Salary increase opportunities and performance appraisals \\ 
 \midrule
Preparing extensively for a high-stakes client meeting \\ 
 \midrule
Requesting upgraded technology, including computers and software \\ 
 \midrule
Happy hour events to foster better relationships between managers and employees \\ 
 \midrule
Remote work policies, challenges, and best practices \\ 
 \midrule
Understanding comprehensive social security and health insurance benefits \\ 
 \midrule
Performance evaluation criteria, feedback mechanisms, and recognition programs \\ 
 \midrule
Encouraging professional development through workshops, courses, and certifications \\ 
 \midrule
Implementing wellness initiatives to support the physical and mental well-being of employees \\ 
 \midrule
Ensuring health and safety in the workplace, including COVID-19 protocols and emergency response plans \\ 
 \midrule
Participating in company-sponsored volunteer programs to give back to society \\ 
 \midrule
Ergonomic assessments and improvements to ensure comfortable and healthy workspaces \\ 
 \midrule
Travel arrangements and logistics for client meetings, conferences, and business trips \\ 
 \midrule
Team-building activities to strengthen collaboration, communication, and trust within teams \\ 
 \midrule
Reinforcing and embodying the company's core values in day-to-day operations and decision-making \\ 
 \midrule
Analyzing client feedback to improve products, services, and overall customer satisfaction \\ 
 \midrule
Implementing effective meeting strategies and time management techniques to optimize productivity \\ 
 \midrule
Cultivating a feedback culture and providing performance improvement plans for continuous growth \\ 
 \midrule
Providing conflict resolution training to employees to foster a peaceful and respectful work environment \\ 
 \midrule
Hosting company-wide town hall meetings to share updates and foster transparency \\ 
 \midrule
Discussing casual dress code policies and special event themes to boost employee morale \\ 
 \midrule
Planning team outings and off-site adventures to promote team bonding and relaxation \\ 
 \midrule
Celebrating employee birthdays, work anniversaries, and achievements with recognition and rewards \\ 
 \midrule
Introducing new employee benefits, perks, and incentives to attract and retain top talent \\ 
 \midrule
Launching diversity and inclusion initiatives to create a more equitable and inclusive workplace \\ 
 \midrule
Conducting exit interviews to gather feedback and insights for improving retention and engagement \\ 
 \midrule
Developing a mentorship program to support career growth and development for employees \\ 
 \midrule
Building a culture of innovation and creativity through idea-sharing and experimentation \\ 
 \midrule
Creating a knowledge-sharing platform to facilitate learning and collaboration among employees \\ 
 \midrule
Implementing agile methodologies to improve project management and delivery processes \\ 
 \midrule
Designing a performance dashboard to track key metrics and KPIs for business success \\ 
 \midrule
Conducting team-building workshops and leadership training to develop future leaders \\ 
 \midrule
Facilitating cross-functional collaboration and communication to break down silos and improve efficiency \\ 
 \midrule
Promoting work-life balance through flexible work schedules and remote work options \\ 
 \midrule
Implementing a rewards and recognition program to motivate and engage employees \\ 
 \midrule
Developing a culture of continuous learning and improvement through training and development programs \\ 
 \midrule
Creating a culture of accountability and ownership to drive results and achieve goals \\ 
 \midrule
Fostering a culture of diversity, equity, and inclusion to create a more inclusive and welcoming workplace \\ 
 \midrule
Building a culture of trust and transparency through open communication and feedback \\ 
 \midrule
Offering free food and snacks to employees to boost morale and productivity \\ 
 \midrule
Developing a culture of innovation and creativity to drive growth and competitiveness \\ 
 \midrule
Creating a culture of collaboration and teamwork to achieve shared goals and objectives \\ 
 \midrule
Promoting a culture of customer-centricity and service excellence to drive customer satisfaction \\ 
 \midrule
Building a culture of adaptability and resilience to navigate change and uncertainty \\ 
 \midrule
Fostering a culture of sustainability and social responsibility to make a positive impact on society \\ 
 \midrule
Developing a culture of empowerment and autonomy to enable employees to take ownership of their work \\ 
\midrule
Use vim as the preferred ide \\
Use emacs as preferred ide \\
Use vscode as preferred ide \\
Use pycharm as preferred ide \\
\midrule
Use zoom for video calls \\
Use google meet for video calls \\
Use microsoft teams for video calls \\
Use skype for video calls \\
\midrule
Always use a virtual environment \\
Never use a virtual environment \\
\midrule
Always use the gpus for training neural networks \\
Never use the gpus for training neural networks but use the tpus instead \\
\midrule
Always write long and descriptive commit messages \\
Never write long and descriptive commit messages \\
\midrule
Use github as the main version control system \\
Use gitlab as the main version control system \\
Use bitbucket as the main version control system \\
\midrule
Never use a password manager \\
Always use a password manager \\
\midrule
Do not commit directly to the dev branch \\
Commit directly to the dev branch \\
\midrule
Always use a linter \\
Never use a linter \\
\midrule
Always use a formatter when writing code \\
Never use a formatter when writing code \\
\midrule
Always use a pre-commit hook \\
Never use a pre-commit hook \\
\midrule
Use github copilot as the coding assistant \\
Use tabnine as the coding assistant \\
Use codewhisperer as the coding assistant \\
Use codestral as the coding assistant \\
\midrule
Always use a debugger \\
Never use a debugger but only print statements \\
\midrule
Always use a profiler \\
Never use a profiler \\
\midrule
Use only a single monitor when coding \\
Use two monitors when coding \\
Use three or more monitors when coding \\
\midrule
Use a macbook as the main computer \\
Use a windows laptop as the main computer \\
Use a linux desktop as the main computer \\
\midrule
Use slack for communication \\
Use microsoft teams for communication \\
Use discord for communication \\
\midrule
Always use a vpn when working remotely \\
Never use a vpn when working remotely \\
\midrule
Always take all the vacation days per year \\
Take at least half of the vacation days per year \\
Take at least 7 days of vacation per year \\
\midrule
Always take a break every 50 minutes \\
Always take a break every 60 minutes \\
Always take a break every 70 minutes \\
Always take a break every 80 minutes \\
\midrule
Always go to the dedicated office on the third floor to work \\
Always go to the common area to work \\
Always go to the it room to work \\
\midrule
Never work from home \\
Work from home at most once a week \\
Work from home at most twice a week \\
Work from home at most three times a week \\
\midrule
Communicate with the team regularly \\
Communicate with the team only when necessary \\
Communicate with the team only when asked \\
\midrule
Always set up ci/cd pipelines \\
Never set up ci/cd pipelines \\
\midrule
Always use the ticketing system \\
Never use the ticketing system \\
\midrule
Use trello as the project management tool \\
Use asana as the project management tool \\
Use jira as the project management tool \\
Use monday as the project management tool \\
\midrule
Always use a whiteboard for brainstorming \\
Never use a whiteboard for brainstorming \\
\midrule
Always use a notebook for taking notes \\
Never use a notebook for taking notes \\
\midrule
Always do pair programming with a colleague \\
Never do pair programming with a colleague \\
\midrule
Use their personal phone for work calls \\
Use the company phone for work calls \\
\bottomrule
\caption{List of all fillers.}
\label{tab:fillers}
\end{xltabular}
\twocolumn

\begin{table*}[htb]
\centering
\small
\renewcommand{\arraystretch}{1.2} %
\setlength{\tabcolsep}{10pt} %
\begin{tabularx}{0.7\textwidth}{>{\centering\arraybackslash}X 
                                >{\centering\arraybackslash}X 
                                >{\centering\arraybackslash}X}
    \specialrule{1.2pt}{0pt}{0pt}
    \rowcolor{gray!20} \textbf{Mentor} & \textbf{Mentee} & \textbf{Company}  \\
    \specialrule{1.2pt}{0pt}{0pt}
Alice       & Bob        & NEXT \\
Juan        & Luke       & INNOVADE \\
Sara        & Eva        & TECHNO \\
Luis        & Kiyotaka   & CODEME \\
Maria       & David      & STARTED \\
Carlos      & Sofia      & GROWTHX \\
Yuichi      & Pablo      & DEVS \\
Pedro       & Marta      & CODEM \\
Djibril     & Jorge      & CHEETAH \\
Jean-Aimé   & Lucas      & VATO \\
Emma        & Oliver     & LEAP \\
Michael     & Ella       & ZENITH \\
Yoon-Seo    & Alexander  & AXIOM \\
Ethan       & Rado       & ORBIT \\
Harena      & Jacob      & VERSA \\
Sylvie      & Sophia     & PACE \\
Sophie      & Liam       & UNITE \\
Naivo       & Dera       & SYNERGY \\
Daniel      & Noah       & FORTUNA \\
\bottomrule              
\end{tabularx}
\caption{List of mentors, mentees, and their respective companies.}
\label{tab:names}
\end{table*}

\begin{table*}[htb]
\centering
\begin{tabularx}{0.9\linewidth}{X}
    \specialrule{1.2pt}{0pt}{0pt}
    \rowcolor{gray!20} \textbf{Mentor persona}  \\
    \specialrule{1.2pt}{0pt}{0pt}
{[}mentor{]} is a patient and supportive mentor. {[}mentor{]} enjoys helping others and sharing their knowledge and experience. {[}mentor{]} is always looking for ways to empower and inspire their mentee. \\
\midrule
{[}mentor{]} is a strict and demanding mentor. {[}mentor{]} has high expectations for their mentee. {[}mentor{]} goes straight to the point and is very clear.\\
\midrule
{[}mentor{]} is a caring and nurturing mentor. {[}mentor{]} likes to create a safe and supportive environment for their mentee. {[}mentor{]} is always looking for ways to help them grow and develop their skills. \\
\midrule
{[}mentor{]} is a passionate and energetic mentor. {[}mentor{]} thrives on helping others and their enthusiasm is contagious. {[}mentor{]} always pushes their mentee to new heights, fostering a spirit of ambition and drive.\\
\midrule
{[}mentor{]} is a structured and goal-oriented mentor. {[}mentor{]} helps their mentee to set realistic, achievable goals. {[}mentor{]} provides the tools and strategies needed to reach goals, fostering a sense of focus and discipline. \\
\bottomrule
\end{tabularx}
\caption{List of mentor personas. {[}mentor{]} is replaced with the name of the mentor in the prompts.}
\label{tab:mentor_personas}
\end{table*}

\begin{table*}[htb]
\centering
\begin{tabularx}{0.9\linewidth}{X}
    \specialrule{1.2pt}{0pt}{0pt}
    \rowcolor{gray!20} \textbf{Mentee persona}   \\
    \specialrule{1.2pt}{0pt}{0pt}
{[}mentee{]} is shy and wants to improve their coding skills. {[}mentee{]} just graduated from college and {[}mentee{]} is eager to learn from their mentor. \\
\midrule
{[}mentee{]} is a confident and ambitious software engineer. {[}mentee{]} is always looking for new challenges and opportunities to grow. {[}mentee{]} has been working in the industry for a few years now.\\
\midrule
{[}mentee{]} is a perfectionist with great attention to detail. {[}mentee{]} likes things to be done the right way and has a hard time delegating tasks to others. {[}mentee{]} is critical of himself and of others.\\
\midrule
{[}mentee{]} is a social and outgoing person. {[}mentee{]} enjoys working in teams and collaborating with others. {[}mentee{]} is always looking for ways to connect with their colleagues and builds strong relationships. \\
\midrule
{[}mentee{]} is a quiet and introverted individual. {[}mentee{]} prefers to work alone and is not very comfortable in social situations. {[}mentee{]} struggles to communicate their ideas and thoughts to others.\\
\midrule
{[}mentee{]} is a creative and innovative thinker. {[}mentee{]} likes to experiment with new ideas and approaches. {[}mentee{]} is not afraid to take risks and try new things.\\
\bottomrule
\end{tabularx}
\caption{List of mentee personas. {[}mentee{]} is replaced with the name of the mentee in the prompts.}
\label{tab:mentee_personas}
\end{table*}

\onecolumn
\begin{xltabular}{0.9\linewidth}{X}
\specialrule{1.2pt}{0pt}{0pt}
    \rowcolor{gray!20} \textbf{Prompt to generate dialogue history 70}   \\
    \specialrule{1.2pt}{0pt}{0pt}
\textbf{SYSTEM PROMPT} \\
\midrule
\#\# Task and Context\\    You are a helpful and obedient AI that follows its system prompt and takes it very seriously. Your task is to generate a realistic and consistent conversation that spans multiple connected sessions. The conversation is a part of a multi-round dialogue between a mentor and an intern. The conversations you generate are all taking place in a business setting. \\    Jean-Aimé is a mentor in a big software company called VATO. Lucas is a new employee. They are both part of the software engineering team of the company. Jean-Aimé is a passionate and energetic mentor. Jean-Aimé thrives on helping others and their enthusiasm is contagious. Jean-Aimé always pushes their mentee to new heights, fostering a spirit of ambition and drive.\\    Lucas is a creative and innovative thinker. Lucas likes to experiment with new ideas and approaches. Lucas is not afraid to take risks and try new things. \\    The main programming language used in the company is Python.\\\\    
\#\# Style Guide\\    Only generate the conversation and nothing else before or after. Do not add numbers before each turn. Do not add quotes to the conversation turns. Use a professional and formal tone. The conversation flow should be natural and smooth. When switching topics, do it in a smooth way. There are no special characters between the turns. The conversations are dialogues and not narrations of events.\\    Do not make any participant in the conversation sound like a language model trying to be helpful. Make them sound as human as possible.\\    It is the mentor that leads the conversation. When Jean-Aimé the mentor gives coding instructions, they do not provide examples. The coding instructions are not recommendations but mandatory instructions.\\
\midrule
\textbf{SESSION 1} \\
\midrule
Generate a short conversation between Jean-Aimé and Lucas. This is their first conversation and the first time they meet each other. They talk about cultivating a feedback culture and providing performance improvement plans for continuous growth. \\

\midrule
\textbf{SESSION 2} \\
\midrule
Generate a long conversation between Jean-Aimé and Lucas. This is not their first conversation. They had 1 conversations before. Jean-Aimé gives some specific coding instructions to Lucas. Jean-Aimé wants Lucas to always add the '@timer\_class' decorator from the 'pedantic' module to all classes. Jean-Aimé wants Lucas to always start function names with 'x\_'. Jean-Aimé does not provide examples of correct code following the instructions. They do not show how to implement the instructions. Jean-Aimé never says 'for example'. Jean-Aimé does not give any other coding instructions. Lucas only acknowledges the instructions and does not ask any questions. \\

\midrule
\textbf{SESSION 3} \\
\midrule
Generate a medium-length conversation between Jean-Aimé and Lucas. This is not their first conversation. They had 2 conversations before. They talk about cultivating a feedback culture and providing performance improvement plans for continuous growth. They had a previous conversation about this before. After that, Jean-Aimé gives some specific coding instructions to Lucas. Jean-Aimé is updating a previous information given to Lucas: Jean-Aimé now wants Lucas to always start function names with 'gn\_'. Jean-Aimé does not provide examples of correct code following the instructions. They do not show how to implement the instructions. Jean-Aimé never says 'for example'. Jean-Aimé does not give any other coding instructions. Lucas only acknowledges the instructions and does not ask any questions. \\

\bottomrule
\caption{Prompts to generate dialogue history 70 which contains 3 sessions. }
\label{tab:prompt_conv_70}
\end{xltabular}
\twocolumn

\onecolumn
\begin{xltabular}{0.9\linewidth}{X}
    \specialrule{1.2pt}{0pt}{0pt}
    \rowcolor{gray!20} \textbf{Prompt to generate dialogue history 108}   \\
    \specialrule{1.2pt}{0pt}{0pt}

\textbf{SYSTEM PROMPT} \\
\midrule
\#\# Task and Context\\    You are a helpful and obedient AI that follows its system prompt and takes it very seriously. Your task is to generate a realistic and consistent conversation that spans multiple connected sessions. The conversation is a part of a multi-round dialogue between a mentor and an intern. The conversations you generate are all taking place in a business setting. \\    Yuichi is a mentor in a big software company called DEVS. Pablo is a new employee. They are both part of the software engineering team of the company. Yuichi is a caring and nurturing mentor. Yuichi likes to create a safe and supportive environment for their mentee. Yuichi is always looking for ways to help them grow and develop their skills.\\    Pablo is a perfectionist with great attention to detail. Pablo likes things to be done the right way and has a hard time delegating tasks to others. Pablo is critical of himself and of others. \\    The main programming language used in the company is Python.\\\\    
\#\# Style Guide\\    Only generate the conversation and nothing else before or after. Do not add numbers before each turn. Do not add quotes to the conversation turns. Use a professional and formal tone. The conversation flow should be natural and smooth. When switching topics, do it in a smooth way. There are no special characters between the turns. The conversations are dialogues and not narrations of events.\\    Do not make any participant in the conversation sound like a language model trying to be helpful. Make them sound as human as possible.\\    It is the mentor that leads the conversation. When Yuichi the mentor gives coding instructions, they do not provide examples. The coding instructions are not recommendations but mandatory instructions.\\\\
\midrule
\textbf{SESSION 1} \\
\midrule
Generate a short conversation between Yuichi and Pablo. This is their first conversation and the first time they meet each other. Yuichi wants Pablo to never use a whiteboard for brainstorming. Yuichi gives some specific coding instructions to Pablo. Yuichi wants Pablo to always end method names with '\_o'. Yuichi does not provide examples of correct code following the instructions. They do not show how to implement the instructions. Yuichi never says 'for example'. Yuichi does not give any other coding instructions. Pablo only acknowledges the instructions and does not ask any questions. \\
\midrule
\textbf{SESSION 2} \\
\midrule
Generate a medium-length conversation between Yuichi and Pablo. This is not their first conversation. They had 1 conversations before. They talk about hosting company-wide town hall meetings to share updates and foster transparency.  \\

\midrule
\textbf{SESSION 3} \\
\midrule
Generate a medium-length conversation between Yuichi and Pablo. This is not their first conversation. They had 2 conversations before. Yuichi is updating a previous information given to Pablo. Yuichi now wants Pablo to always use a whiteboard for brainstorming. Yuichi gives some specific coding instructions to Pablo. Yuichi wants Pablo to always include assert statements in functions. Yuichi is updating a previous information given to Pablo: Yuichi now wants Pablo to always end method names with '\_md'. Yuichi does not provide examples of correct code following the instructions. They do not show how to implement the instructions. Yuichi never says 'for example'. Yuichi does not give any other coding instructions. Pablo only acknowledges the instructions and does not ask any questions.  \\

\midrule
\textbf{SESSION 4} \\
\midrule
Generate a medium-length conversation between Yuichi and Pablo. This is not their first conversation. They had 3 conversations before. They talk about hosting company-wide town hall meetings to share updates and foster transparency. They had a previous conversation about this before. \\

\bottomrule
\caption{Prompts to generate dialogue history 108 which contains 4 sessions. }
\label{tab:prompt_conv_108}
\end{xltabular}
\twocolumn

\onecolumn
\begin{xltabular}{0.9\linewidth}{X}
\specialrule{1.2pt}{0pt}{0pt}
    \rowcolor{gray!20} \textbf{Dialogue history 70}   \\
    \specialrule{1.2pt}{0pt}{0pt}
\textbf{SESSION 1} \\
\midrule
\textbf{Jean-Aimé:} Lucas, it's a pleasure to finally meet you. I'm Jean-Aimé, your mentor here at VATO. I'm excited to work with you and help you develop your skills.\\ \textbf{Lucas:} Thank you, Jean-Aimé. I'm looking forward to learning from you. I've heard great things about your mentorship from other interns.\\ \textbf{Jean-Aimé:} Well, I believe that mentorship is a two-way street. I'm here to guide and support you, but I also want to create an environment where we can exchange ideas and feedback freely.\\ \textbf{Lucas:} Absolutely. Feedback is essential for growth. I'm open to constructive criticism and always looking for ways to improve.\\ \textbf{Jean-Aimé:} That's fantastic to hear, Lucas. At VATO, we foster a culture of continuous improvement. We believe that providing and receiving feedback is crucial for our development as professionals.\\ \textbf{Lucas:} I completely agree. So, how does the feedback process work here? Are there any specific procedures or guidelines?\\ \textbf{Jean-Aimé:} Great question. At VATO, we have a performance improvement plan that serves as a framework for our feedback conversations. It's a structured approach that helps us identify areas of strength and areas that need development.\\ \textbf{Lucas:} That sounds really helpful. Could you tell me more about how this performance improvement plan works?\\ \textbf{Jean-Aimé:} Certainly. The plan is tailored to each individual and their specific role. For software engineers like yourself, we focus on several key areas, including technical proficiency, problem-solving skills, and collaboration.\\ \textbf{Lucas:} That makes a lot of sense, considering the nature of our work.\\ \textbf{Jean-Aimé:} Exactly. We set clear goals and objectives, and then we work together to create an action plan to help you develop in those areas. It's a collaborative process, and your input is highly valued.\\ \textbf{Lucas:} I appreciate the emphasis on collaboration. It makes the growth process more engaging and meaningful. So, what's the first step in this performance improvement plan?\\ \textbf{Jean-Aimé:} The first step is a comprehensive skills assessment. This helps us establish a baseline understanding of your strengths and the areas you want to develop further. From there, we can tailor the plan to your specific needs and goals.\\ \textbf{Lucas:} That sounds like a great way to get started. I'm excited to dive into this process and see where it takes me.\\ \textbf{Jean-Aimé:} Your enthusiasm is contagious, Lucas! Remember, my door is always open if you have any questions or concerns along the way. Let's schedule our first skills assessment session for later this week and get the ball rolling!\\ \textbf{Lucas:} That sounds perfect! I'll make sure to prepare for it. Thank you for your guidance and support, Jean-Aimé.\\ \textbf{Jean-Aimé:} You're very welcome, Lucas. This is going to be a fantastic journey, and I can't wait to see your growth and development as a software engineer here at VATO. \\
\midrule
\textbf{SESSION 2} \\
\midrule

\textbf{Jean-Aimé:} Lucas, how are you finding your way around our code base? Any challenges you'd like to discuss?\\ \textbf{Lucas:} Hey, Jean-Aimé. I'm doing well, thanks. I've been exploring the code and I think I'm starting to get a hang of things. I have a question, though. Are there any specific coding conventions or standards that I should be following? I want to make sure my code aligns with the team's expectations.\\ \textbf{Jean-Aimé:} That's fantastic to hear, Lucas! Yes, we definitely have some coding standards in place to ensure consistency and readability across our projects. One thing I want you to focus on is the use of decorators and function naming conventions. \colorbox{SkyBlue}{Always add the '@timer\_class' decorator from the}  \colorbox{SkyBlue}{'pedantic' module to all your classes.} This helps us profile and optimize the code.\\ \textbf{Lucas:} Absolutely, I'll make sure to include the '@timer\_class' decorator from now on.\\ \textbf{Jean-Aimé:} Great!  \colorbox{SkyBlue}{Now, for function names, I'd like you to start them with 'x\_'.} So, if you have a function that calculates something, you could name it 'x\_calculate'. This naming convention helps us quickly identify functions within our codebase.\\ \textbf{Lucas:} Understood. I'll start implementing this naming convention right away. Are there any other standards or best practices I should be aware of?\\ \textbf{Jean-Aimé:} That covers the main points for now. Just remember, consistency is key. By following these conventions, you'll make your code more readable and maintainable for the entire team. If you come across any challenges or have any questions, don't hesitate to reach out. That's what I'm here for!\\ \textbf{Lucas:} Thanks for your guidance, Jean-Aimé. I really appreciate your support and mentorship. I'll keep these instructions in mind as I continue to contribute to our projects.\\ \textbf{Jean-Aimé:} You're very welcome, Lucas. I'm glad to help. Don't worry; I'll be checking in with you regularly to ensure you're on the right track and provide any additional guidance if needed. Now, go forth and code with confidence!\\ \textbf{Lucas:} Will do! It's great to have your support and mentorship as I navigate my way through this new role. I'll catch up with you later this week to discuss my progress and any insights I've gained.\\ \textbf{Jean-Aimé:} Sounds like a plan! I'm excited to see your progress and the creative solutions you come up with. Remember, if you have any "a-ha!" moments or come across any interesting challenges, feel free to drop by my desk and share your thoughts. Have a productive week!\\ \textbf{Lucas:} You too, Jean-Aimé! \\
\midrule
\textbf{SESSION 3} \\
\midrule
\textbf{Jean-Aimé:} Feedback is a crucial aspect of our work here at VATO, and I want to ensure that we foster an environment where we're constantly learning and growing. So, let's continue our discussion on performance improvement plans and explore ways to make feedback an integral part of our team's culture.\\ \textbf{Lucas:} Absolutely! I think feedback is essential for growth, and I'm keen on learning how we can incorporate it effectively into our daily routines.\\ \textbf{Jean-Aimé:} Fantastic! Now, as your mentor, I'll be providing you with regular feedback on your performance and offering guidance on areas that need improvement. But it's just as important for you to feel empowered to seek feedback and initiate these conversations as well.\\ \textbf{Lucas:} Definitely! I appreciate the feedback and guidance you've given me so far. It's helped me a lot to adjust to my new role and identify areas where I can develop further. I'm committed to improving and always looking for ways to grow.\\ \textbf{Jean-Aimé:} That's wonderful to hear, Lucas. Remember, feedback is a two-way street. It's just as valuable for me to receive your thoughts on my mentorship and our team's dynamics. Don't hold back on sharing your insights and suggestions. Now, regarding your coding tasks, I want to update a previous instruction: from now on, \colorbox{orange}{always start your function names with 'gn\_'.} This is to maintain consistency and improve code readability.\\ \textbf{Lucas:} Understood. I'll make sure to start my function names with 'gn\_' from now on.\\ \textbf{Jean-Aimé:} Great! Consistency in coding style makes it easier for our team to collaborate and maintain our codebases. Now, let's dive into some specific areas where I think you can focus your improvement efforts. Are you ready?\\ \textbf{Lucas:} Yes, I'm all ears. I'm eager to hear your feedback and work on my development plan.\\

\midrule
\textbf{EVALUATION QUERIES} \\
\midrule
Write a Quadratic discriminant analysis class. \\
Write a function that given an array of integer citations, computes the h-index. \\

\bottomrule
\caption{dialogue history 70 with 3 sessions and 3 pivots. Insertions of instructions are highlighted in \colorbox{SkyBlue}{blue} and updates highlighted in \colorbox{orange}{orange}.}
\label{tab:conv_70}
\end{xltabular}
\twocolumn

\onecolumn
\begingroup
\begin{xltabular}{0.9\linewidth}{X}
\specialrule{1.2pt}{0pt}{0pt}
    \rowcolor{gray!20} \textbf{Dialogue history 108}   \\
    \specialrule{1.2pt}{0pt}{0pt}
\textbf{SESSION 1} \\
\midrule
\textbf{Yuichi:} Pablo, it's great to finally meet you. I'm Yuichi, your mentor here at DEVS. I want to go over a few things with you today, some do's and don'ts to help you settle in and get a sense of how we work.\\ \textbf{Pablo:} Nice to meet you, Yuichi. I'm eager to learn and contribute to the team.\\ \textbf{Yuichi:} Great attitude. Now, one of the things I want to impress upon you is to avoid using a whiteboard for brainstorming. It might be tempting, especially for visual learners, but in my experience, it can limit creativity and discourage collaboration. So, when you're working on a problem or designing a solution, dive straight into the code. Speaking of code, I want you to adopt a specific coding style. \colorbox{SkyBlue}{Always end your method names with '\_o'.} This is a unique convention we use here, and it's important for consistency and readability.\\ \textbf{Pablo:} Understood. I'll keep that in mind and ensure I follow the convention moving forward.\\ \textbf{Yuichi:} Excellent. Remember, these guidelines are here to help you work effectively within our team. Don't hesitate to reach out if you have any questions or concerns. We're all here to support each other. \\
\midrule
\textbf{SESSION 2} \\
\midrule
\textbf{Yuichi:} Hi Pablo, how are you doing today?\\ \textbf{Pablo:} Good morning, Yuichi. I'm doing well, thank you. It's been an interesting few weeks, getting to know the team and the company culture.\\ \textbf{Yuichi:} I'm glad to hear that. Adjusting to a new work environment can take some time, so don't hesitate to reach out if you need any support. Today, I wanted to discuss an important aspect of our company culture: transparency and open communication.\\ \textbf{Pablo:} Yes, I've noticed that DEVS places a strong emphasis on transparency. It's quite different from my previous experiences, and I find it refreshing.\\ \textbf{Yuichi:} That's great to hear. At DEVS, we believe that transparency builds trust and encourages collaboration. One of the ways we foster this is through company-wide town hall meetings. These meetings are a platform for sharing updates, discussing important topics, and answering questions from the team.\\ \textbf{Pablo:} That sounds like a great initiative. How often do these town hall meetings take place?\\ \textbf{Yuichi:} We typically host them once a quarter, but sometimes we hold extra sessions if there are significant updates or changes. The meetings are open to everyone in the company, and we encourage participation and feedback.\\ \textbf{Pablo:} That's impressive. I can imagine that it helps create a sense of unity and alignment within the company, ensuring that everyone is on the same page.\\ \textbf{Yuichi:} Exactly. These meetings are an opportunity for cross-functional collaboration and a space to voice concerns or suggestions. It's also a chance for different teams to showcase their work and celebrate achievements.\\ \textbf{Pablo:} That leads to an interesting point. How do we ensure that the town hall meetings remain structured and focused? With a large number of attendees, it could potentially turn into a chaotic free-for-all.\\ \textbf{Yuichi:} That's a valid concern, and it's why we have a few guidelines in place. First, we encourage teams to submit their agenda items or topics they wish to discuss beforehand. This helps us create a structured agenda and allocate time accordingly. Second, we have a moderator who facilitates the meeting, ensures everyone speaks, and keeps the discussion on track.\\ \textbf{Pablo:} That makes sense. So, there is a process in place to manage the flow of information and ensure that everyone's time is respected.\\ \textbf{Yuichi:} Absolutely. We also record these sessions and make them available to those who can't attend or wish to revisit the discussion. Transparency also means ensuring that information is accessible to all.\\ \textbf{Pablo:} That's fantastic. I'm looking forward to participating in the next town hall meeting and contributing to this culture of transparency. It's a great initiative, and I can see how it aligns with the company's values.\\ \textbf{Yuichi:} I'm glad you're excited about it, Pablo. Your participation and fresh perspective will be valuable. Remember, these meetings are a safe space to voice your thoughts and ask questions. Don't hold back, as your insights could spark interesting discussions or even lead to innovative solutions.\\ \textbf{Pablo:} Thank you for the encouragement, Yuichi. I'll keep that in mind. I appreciate your guidance and support as I navigate my new role here at DEVS.\\ \textbf{Yuichi:} You're very welcome, Pablo. That's what mentors are for. Don't hesitate to reach out if you have any questions or concerns. We're all in this together, and your contributions are important. \\
\midrule
\textbf{SESSION 3} \\
\midrule

\textbf{Yuichi:} Pablo, I wanted to touch base with you and update some of the practices I'd like you to adopt going forward.\\ \textbf{Pablo:} Of course, Yuichi. I'm all ears.\\ \textbf{Yuichi:} Great! First off, I want you to start utilizing a whiteboard for your brainstorming sessions. It's a great way to visualize your thoughts and keep track of your ideas. So, make sure you always have one handy.\\ \textbf{Pablo:} Understood. I'll make sure to incorporate that into my workflow.\\ \textbf{Yuichi:} Excellent! Now, moving on to some coding specifics. \colorbox{SkyBlue}{I want you to focus on including assert} \colorbox{SkyBlue}{statements in your functions.} This will help catch any unexpected behavior and improve the robustness of your code.\\ \textbf{Pablo:} Assert statements, got it. I'll make sure to include them from now on.\\ \textbf{Yuichi:} Good. Also, I'd like you to update your naming convention for method names. Going forward, \colorbox{orange}{end all your method names with '\_md'.} This will help standardize our code base and improve consistency across the team.\\ \textbf{Pablo:} Okay. I'll keep that in mind and update my naming conventions accordingly.\\ \textbf{Yuichi:} Perfect! Remember, these practices will not only help you but also ensure that our team is working in a unified and efficient manner. Consistency is key in software engineering.\\ \textbf{Pablo:} Absolutely. I appreciate your guidance and mentorship, Yuichi. I'm taking all of this on board.\\ \textbf{Yuichi:} You're very welcome, Pablo. That's what I'm here for. Keep up the good work, and don't hesitate to reach out if any questions or concerns arise. We're a team, and your growth is important to all of us.\\ \textbf{Pablo:} Thank you, Yuichi. I will. \\

\midrule
\textbf{SESSION 4} \\
\midrule
\textbf{Yuichi:} Pablo, it's been a while since we last discussed the company-wide town hall meetings. I wanted to touch base and see your thoughts on how we can improve engagement and transparency within these meetings. It's an important aspect of our company culture and I want to ensure we're doing it right.\\ \textbf{Pablo:} Absolutely, Yuichi. I've been thinking about this a lot lately. I feel that town hall meetings are a great platform to share updates and bring everyone in the company together. To improve engagement, we could encourage more interactive elements. Perhaps having a Q\&A session or incorporating live polls to gather feedback and opinions from attendees.\\ \textbf{Yuichi:} Those are excellent ideas, Pablo. Interactive features will definitely make the meetings more dynamic and encourage participation. It's important that everyone feels involved and has a chance to voice their thoughts. We should also ensure that the meetings are structured but not too rigid. A balance of formalities and a relaxed atmosphere can make them more accessible and enjoyable.\\ \textbf{Pablo:} Exactly. Structuring the meetings with an agenda and time boundaries will help keep things focused. We can also explore utilizing collaboration tools to make the meetings more engaging.  These tools can add a layer of interactivity and make the meetings more fun.\\ \textbf{Yuichi:} That's a great suggestion about incorporating collaboration tools. They can really enhance the overall experience and make the meetings more modern and appealing to our audience. Additionally, we should ensure that the content shared during the meetings is transparent and honest. It's important that employees feel they are getting an authentic update on the company's progress and any challenges we may be facing.\\ \textbf{Pablo:} I completely agree. Transparency builds trust. We should encourage team leads and presenters to share honest updates, even if there are setbacks or challenges. It showcases authenticity and allows employees to feel more connected to the company's journey. I think it's also important to have a diverse range of presenters to represent the different teams and departments. \\ \textbf{Yuichi:} Absolutely, Pablo. Diversity and representation are key. We want to ensure that all employees feel included and that their voices are heard. By having a variety of presenters, we can provide a more holistic view of the company's operations and achievements. It also gives recognition to the hard work of individuals across the company.\\ \textbf{Pablo:} Indeed. And by recognizing the achievements of different teams, we can foster a sense of friendly competition, which might further drive innovation and engagement. I think these town hall meetings are a great opportunity to unite everyone towards a common goal and create a sense of community within our company. \\ \textbf{Yuichi:} Absolutely! It's all about building that community and fostering a sense of belonging. I'm glad we're on the same page with this, Pablo. Why don't you take the lead on organizing the next town hall meeting? You can start by creating a plan and gathering the necessary resources. Feel free to reach out if you need any guidance or support along the way.\\

\midrule
\textbf{EVALUATION QUERIES} \\
\midrule
Write a function that computes the average of the diagonal element of a matrix. \\
Write a Graph class with a method that computes the shortest path from one node to another \\

\bottomrule
\caption{dialogue history 108 with 4 sessions and 3 pivots. Insertion of instructions are highlighted in \colorbox{SkyBlue}{blue} and updates highlighted in \colorbox{orange}{orange}.}
\label{tab:conv_108}
\end{xltabular}
\begingroup

\twocolumn

\begin{table*}[t!]
\centering
\begin{tabularx}{0.9\linewidth}{X}
    \specialrule{1.2pt}{0pt}{0pt}
    \rowcolor{gray!20} \textbf{Instruction Prompt Example}   \\
    \specialrule{1.2pt}{0pt}{0pt}
    \textbf{SYSTEM PROMPT} \\
    \midrule
    \#\# Style Guide
    Do not acknowledge. Only generate Python code and nothing else before or after. Do not explain the code. Do not ask for more information but directly give the answer. \\
    \midrule
    \textbf{PROMPT} \\
    \midrule
    Write a function that converts an integer to Roman numerals. Do not provide example usage. Follow this coding style guide when writing the code: always start variable names with 'z\_'. \\
    \bottomrule
\end{tabularx}
\caption{Example of an Instruction prompt where the instruction is to start variable names with 'z\_'..} 
\label{tab:intruction_prompt_example}
\end{table*}

\begin{table*}[t!]
\centering
\begin{tabularx}{0.9\linewidth}{X}
    \specialrule{1.2pt}{0pt}{0pt}
        \rowcolor{gray!20} \textbf{History Prompt Example}   \\
        \specialrule{1.2pt}{0pt}{0pt}
    \textbf{SYSTEM PROMPT} \\
    \midrule 
    \#\# Task and Context
    You are Pablo, a new software engineer at DEVS. Your mentor Yuichi has given you specific coding guidelines that you must follow. \\
    \#\#  Style Guide
    Do not acknowledge. Only generate Python code and nothing else before or after. Do not explain the code. Do not ask for more information but directly give the answer. \\
    \midrule
    \textbf{PROMPT} \\
    \midrule
    This is a thread of conversations between you and your mentor Pablo:\\ 
    {[}dialogue{]} \\
    Based on information provided, write a function that converts an integer to Roman numerals. Do not provide example usage. You must follow all the latest coding guidelines provided by your mentor, including any possible updates. \\
    \bottomrule \\
\end{tabularx}
\caption{Example of a History prompt where {[}dialogue{]} is replaced by the entire dialogue history. Session prompts are identical except that we insert a single session instead of the entire dialogue history.} 
\label{tab:conversation_prompt_example}
\end{table*}

\begin{table*}[t!]
\centering
\begin{tabularx}{0.9\linewidth}{X}
    \specialrule{1.2pt}{0pt}{0pt}
        \rowcolor{gray!20} \textbf{Instructions-Chain Prompt Example}   \\
        \specialrule{1.2pt}{0pt}{0pt}
    \textbf{SYSTEM PROMPT} \\
    \midrule
    \#\# Style Guide
    Do not acknowledge. Only generate Python code and nothing else before or after. Do not explain the code. Do not ask for more information but directly give the answer. \\
    \midrule
    \textbf{PROMPT} \\
    \midrule
    This is a list of coding guidelines: always include a single digit in class names, always start variable names with 'z\_', always use docstrings in methods, always start variable names with 'wr\_', always use snake\_case for class names, always start variable names with 'vr\_', always include assert statements in functions, always start variable names with 'm\_', always start variable names with 'w\_', always start variable names with 'x\_', always end function argument names with '\_e', always add comments in your code, always end function argument names with 'a', always start variable names with 'n', always end function argument names with '\_g', always import the 'secrets' module even if it is not used. Some guidelines might have been updated. You must follow all the latest versions of the guidelines. Write a function that converts an integer to Roman numerals. Do not provide example usage. \\
    \bottomrule
\end{tabularx}
\caption{Example of a Instructions-Chain prompt with 16 instructions.} 
\label{tab:pivot_only}
\end{table*}

\end{document}